\documentclass{article}

\usepackage{arxiv}
\usepackage{multirow}
\usepackage{setspace}
\usepackage[normalem]{ulem}
\usepackage[utf8]{inputenc} % allow utf-8 input
\usepackage[T1]{fontenc}    % use 8-bit T1 fonts
\usepackage{hyperref}       % hyperlinks
\usepackage{url}            % simple URL typesetting
\usepackage{booktabs}       % professional-quality tables
\usepackage{amsfonts}       % blackboard math symbols
\usepackage{nicefrac}       % compact symbols for 1/2, etc.
\usepackage{microtype}      % microtypography
\usepackage{xcolor}         % colors
\usepackage{todonotes}

\newtheorem{thm}{Theorem}
\newtheorem{lem}{Lemma}
\newtheorem{assumption}{Assumption}

\newtheorem{corollary}{Corollary}
\newtheorem{remark}{Remark}

\newcommand{\bc}[1]{{\color{blue} #1}}
\usepackage{bm}
\DeclareMathAlphabet{\pazocal}{OMS}{zplm}{m}{n}
\newcommand{\Lb}{\pazocal{L}}
\newcommand{\thetab}{\bm{\theta}}

\newcommand\bo[1]{\textbf{ \normalfont{#1}}}
\usepackage{mathtools}
\DeclarePairedDelimiterX{\inp}[2]{\langle}{\rangle}{#1, #2}
\usepackage{tikz}
\newcommand*\circled[1]{\tikz[baseline=(char.base)]{\node[shape=circle,draw,inner sep=2pt] (char) {#1};}}
\newcommand{\norm}[1]{\left\lVert#1\right\rVert}
\usepackage{caption}
\usepackage{subcaption}

\title{TENGraD: Time-Efficient Natural Gradient Descent with Exact Fisher-block Inversion
}
\author{%
 Saeed Soori \\
 Department of Computer Science\\
University of Toronto\\
 \texttt{saeed.soori.sh@gmail.com} \\
 \And
Bugra Can\\
Department of MSIS\\
Rutgers Business School\\
\texttt{bugra.can@rutgers.edu}\\
\And
Baourun Mu \\
   Department of Computer Science \\
   University of Toronto\\
\texttt{baorun.mu@mail.utoronto.ca }\\
\And
   Mert G\"urb\"uzbalaban \\
   Department of MSIS \\
   Rutgers Business School\\
   \texttt{mert.gurbuzbalaban@rutgers.edu} 
   \And
   Maryam Mehri Dehnavi \\
   Department of Computer Science\\
   University of Toronto \\
   \texttt{mmehride@cs.toronto.edu} \\
}

\begin{document}
\date{}
\maketitle

\begin{abstract}
This work proposes a time-efficient Natural Gradient Descent method, called TENGraD, with linear convergence guarantees. Computing the inverse of the neural network’s Fisher information matrix is expensive in NGD because the Fisher matrix is large. Approximate NGD methods such as KFAC attempt to improve NGD's running time and practical application by reducing the Fisher matrix inversion cost with approximation. However, the approximations do not reduce the overall time significantly and lead to less accurate parameter updates and loss of curvature information. TENGraD, improves the time efficiency of NGD by computing Fisher-block inverses with  a computationally efficient  covariance factorization and reuse method. It computes the inverse of each block exactly using the Woodbury matrix identity to preserve curvature information while admitting  (linear) fast convergence rates. Our experiments on image classification tasks for state-of-the-art deep neural architecture on  CIFAR-10, CIFAR-100, and Fashion-MNIST show that TENGraD significantly outperforms state-of-the-art NGD methods and often stochastic gradient descent in wall-clock time.\looseness=-1

\end{abstract}
% \keywords{Deep learning  \and second-order  \and Fisher information matrix }

\section{Introduction}
 Second order methods specifically Natural Gradient Descent (NGD) \cite{amari1998natural, NIPS1996_39e4973b, cai2019gram, NEURIPS2020_7b41bfa5} have gained traction in recent years as they accelerate 
 %the convergence rate of 
 the training of deep neural networks (DNN) %training 
 by capturing the geometry of the
optimization landscape \cite{martens2020new} with the Fisher Information Matrix (FIM) \cite{NEURIPS2018_18bb68e2, NEURIPS2020_7b41bfa5,karakida2019universal}. NGD's running time depends on both the convergence rate and computations per iteration which typically involves computing and inverting the FIM \cite{grosse2016kronecker}. Exact NGD \cite{NEURIPS2019_1da546f2} methods demonstrate improved convergence compared to first order techniques such as Stochastic Gradient Descent (SGD)\cite{robbins1951stochastic}, and approximate NGD variants attempt to improve the time per iteration, hence the overall time, with approximation \cite{grosse2016kronecker, NEURIPS2018_48000647, NEURIPS2020_192fc044}. However, to our knowledge, none of the current approximate NGD approaches outperform or are comparable with the end-to-end wall-clock time of tuned SGD when training DNNs \cite{grosse2016kronecker, NEURIPS2018_48000647, NEURIPS2020_192fc044}.

Exact NGD methods such as \cite{NEURIPS2019_1da546f2} and \cite{bernacchia2019exact} improve the convergence  of training DNNs compared to first-order methods, however, they are typically expensive  and do not scale due to high computational complexity per iteration. The work in \cite{bernacchia2019exact} derives an analytic formula for the linear network using the generalized inverse of the curvature which cannot be used for training state-of-the-art models due to lack of non-linear activation functions.
% . Their method % in \cite{bernacchia2019exact} is only applicable to linear networks  and therefore
Zhang \textit{et al.} \cite{NEURIPS2019_1da546f2}  extend NGD to deep nonlinear networks with non-smooth activations and show that NGD converges to the global optimum with a linear rate. However, their method fails to scale to large or even moderate size models primarily because it relies heavily on backpropagating Jacobian matrices, which scales with the network's output dimension \cite{dangel2019backpack}. In \cite{ren2019efficient}  the authors use Woodbury identity for the inversion of Fisher matrix and propose a unified framework for subsampled Gauss-Newton and NGD methods. Their  framework is limited to fully connected networks and  relies on empirical Fisher and requires extra forward-backward passes to perform parameter updates which slows down the training \cite{NEURIPS2019_46a558d9, dangel2019backpack}.

Approximate NGD approaches such as  \cite{grosse2016kronecker, NEURIPS2018_48000647, NEURIPS2020_192fc044, botev2017practical, heskes2000natural, NIPS2015_2de5d166} attempt to improve the overall execution time of NGD with  FIM  inverse approximation, however, due to costly operations in the inverse approximation process they do not noticeably reduce the overall NGD time. In these methods, the FIM is approximated with a block-diagonal matrix, where blocks correspond to layers. The dimension of each block scales with the input and output dimension of the layer and therefore it cannot be computed and inverted  efficiently  for wide layers. To alleviate this cost, some methods further approximate %is applied 
each block inverse to reduce its size and computation complexity. For example, KFAC \cite{grosse2016kronecker},  approximates each block inverse using the Kronecker product of two smaller matrices, i.e. Kronecker factors. However, these factors have large sizes for wide layers and hence their inversion is expensive. EKFAC  \cite{NEURIPS2018_48000647} improves the approximation used in KFAC by rescaling the Kronecker factors with a diagonal matrix obtained via costly singular value decompositions. Other work
such as KBFGS \cite{NEURIPS2020_192fc044} further estimates the inverse of Kronecker factors using low-rank BFGS types updates. WoodFisher  \cite{singh2020woodfisher} estimates the empirical FIM block inverses using rank-one updates, however, this estimation will not contain enough useful curvature information to produce a good search direction \cite{martens2015optimizing}. Our proposed method, TENGraD, improves the time efficiency of NGD by reducing the FIM block inversion cost using a computationally efficient covariance factorization and reuse of intermediate values method.

Recent work such as \cite{NEURIPS2019_1da546f2,bernacchia2019exact} show that exact NGD converges to the global optimum with a linear convergence rate. Work such as \cite{ren2019efficient} analyze convergence  for  the Levenberg-Marquardt variant of  NGD without specifying the convergence rate. %However, there is limited work on convergence analysis of approximate NGD methods.
The work in \cite{NEURIPS2019_1da546f2} provides the convergence analysis of KFAC for a shallow two-layer fully connected network where the convergence rate depends on the condition number of input data. Goldfarb \textit{et al.} \cite{NEURIPS2020_192fc044} follow  the framework for stochastic quasi-Newton methods and prove that KBFGS converges with a sublinear rate for a network with bounded activation functions. TENGraD improves the convergence properties of approximate NGD methods and  has a linear convergence rate  for DNNs with non-smooth activation functions.

\begin{figure}[h!]
     \centering
     \begin{subfigure}[ht]{0.32\textwidth}
         \centering
         \includegraphics[width=\textwidth]{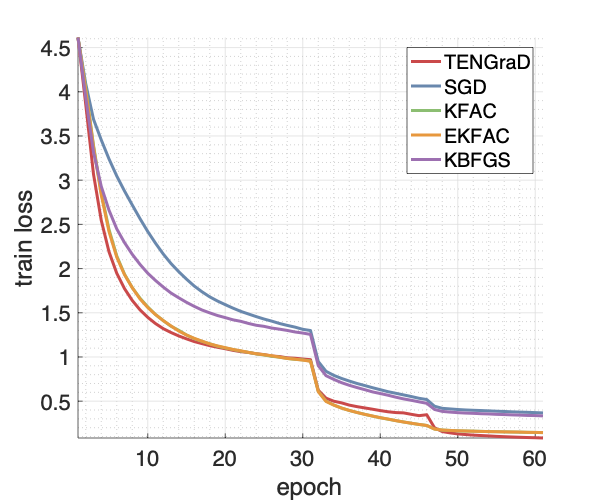}
         \caption{Loss vs epochs.}
         \label{fig:c100-densenet-epoch}
     \end{subfigure}
     \hfill
     \begin{subfigure}[ht]{0.32\textwidth}
         \centering
         \includegraphics[width=\textwidth]{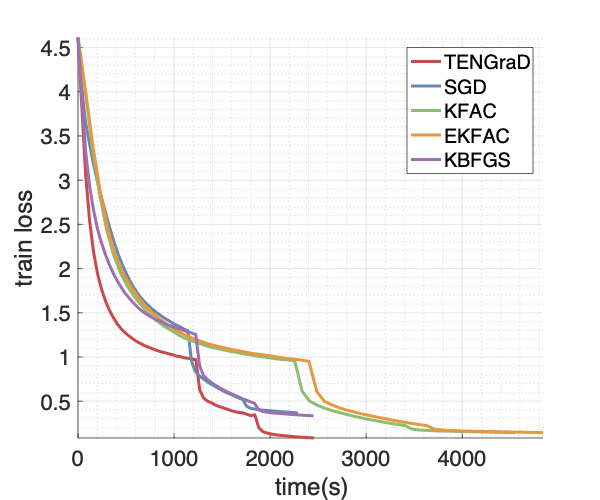}
         \caption{Loss vs time.}
         \label{fig:c100-densenet-time}
     \end{subfigure}
     \hfill
     \begin{subfigure}[ht]{0.32\textwidth}
         \centering
         \includegraphics[width=\textwidth]{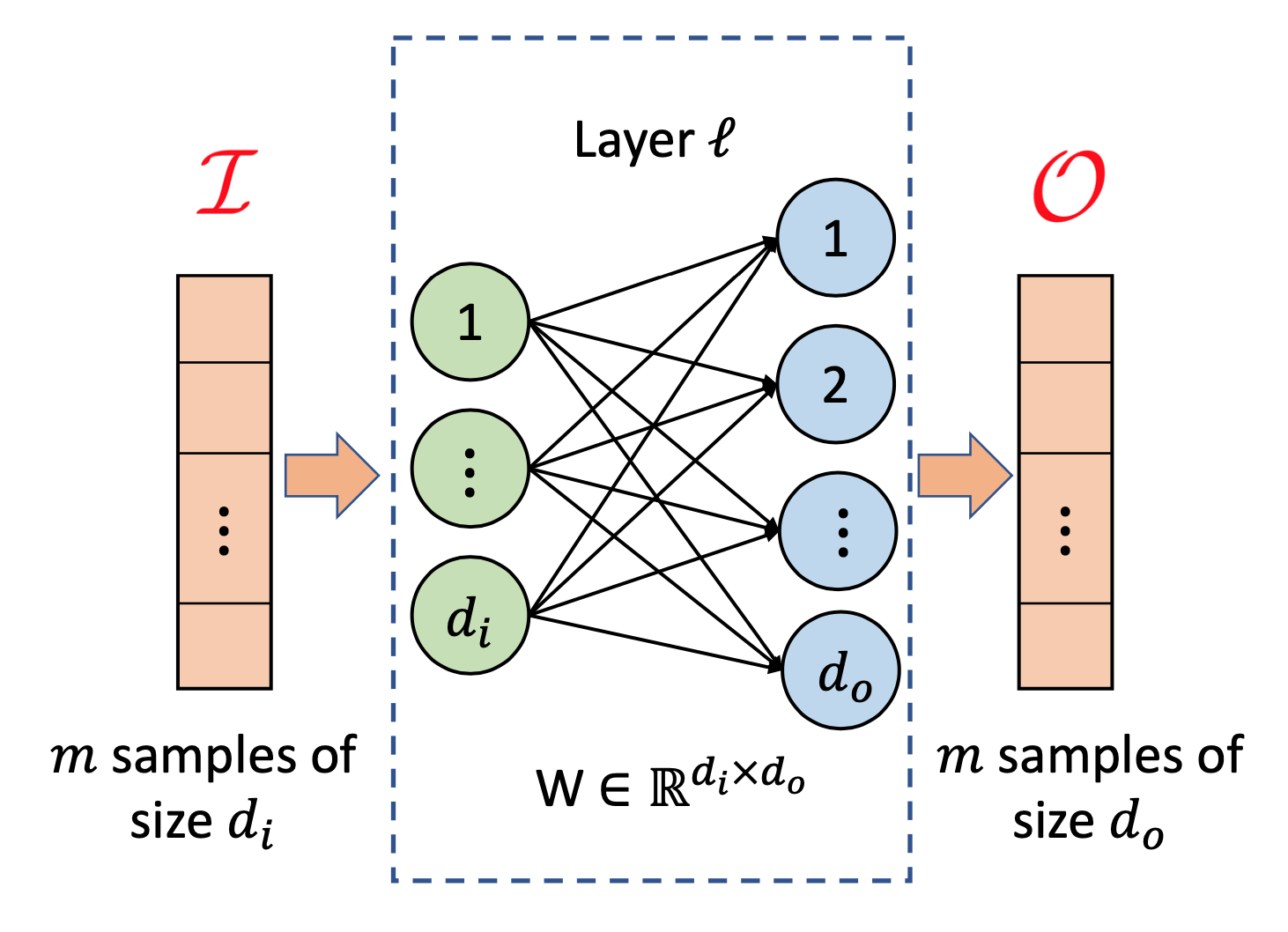}
         \caption{Fully connected layer.}
         \label{fig:main_fc}
     \end{subfigure}
        \caption{ Training loss vs epoch and time for CIFAR-100 on DenseNet, and a fully connected layer.
        \label{fig:motivation}}
\end{figure}

\textbf{Motivation and contributions}: 
\autoref{fig:c100-densenet-epoch} and \autoref{fig:c100-densenet-time} show the performance of approximate NGD methods (including TENGraD) and SGD for an image classification benchmark on a DNN. As shown in \autoref{fig:c100-densenet-epoch},  the NGD methods provide more accurate updates resulting in a better loss reduction compared to SGD. However, as shown in \autoref{fig:c100-densenet-time}, despite efforts made by approximate NGD methods to improve the overall time, none  are competitive with tuned SGD with momentum. TENGraD is the first NGD based implementation, which we call a time-efficient implementation, that outperforms other approximate NGD methods as well as SGD (under certain conditions) in overall time while benefiting from the convergence properties of second-order methods.

\begin{figure}
     \centering
         \includegraphics[width=.85\textwidth]{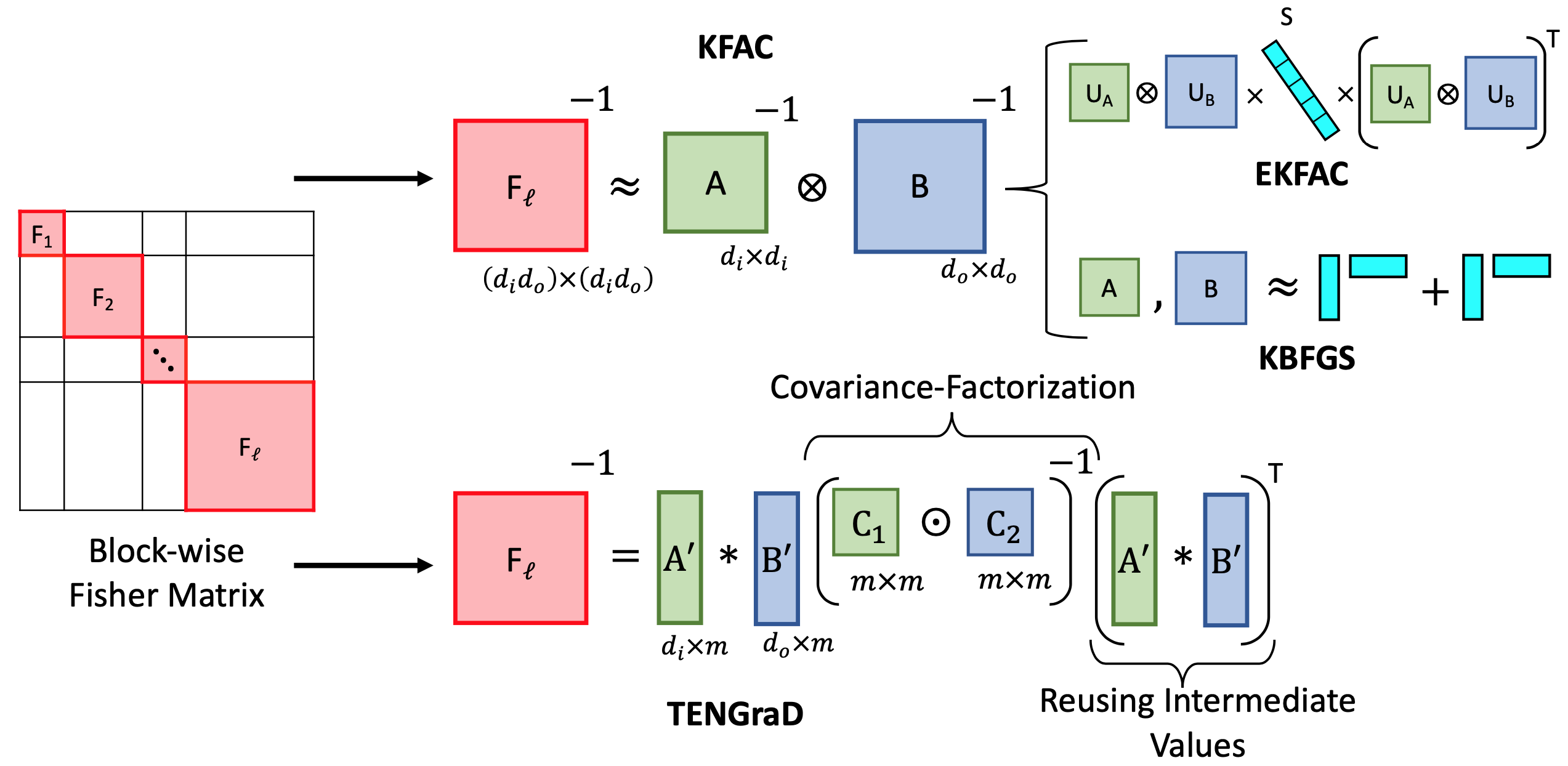}
         \caption{The TENGraD method vs Kronecker-factored approaches for a fully connected layer. %TENGraD efficiently computes the exact Fisher-block inverse while Kronecker approaches perform approximations.
         }
     \label{fig:main_ideas}
\end{figure}

 In the following, we demonstrate for a fully connected layer shown in Figure 1c, %using \autoref{fig:main_ideas},  
 why TENGraD is more time-efficient compared to state-of-the-art approximate NGD methods. % and point to its fast convergence guarantees under certain assumptions. 
 \autoref{fig:main_ideas} shows  KFAC, EKFAC, KBFGS, and TENGraD. All  methods first use a block-approximate approach to create the blocked matrix $F_l$. KFAC, EKFAC, and KBFGS  approximate $F_l$  with a Kronecker product of two Kronecker factors $A$ and $B$. Because the size of factors is proportional to the input and output dimension of a layer, i.e. $d_i$ and $d_o$, their inversion is costly to compute for wide layers and is cubic in input and output dimensions, i.e. $\mathcal{O}(d_i^3 + d_o^3)$.EKFAC creates factors $U_A$ and $U_B$ with a complexity that is also cubic in input and output dimensions of a layer. It also computes the scaling factor $S$ by performing $m$ number of matrix multiplications with the SVD factors. 
 The rank-2 BFGS updates in KBFGS  %to estimate the inverse of the factors  which
 lead to 
 %to a method with 
 a computation complexity that is quadratic in the input and output size of a layer.
 
 TENGraD  computes the exact inverse of Fisher-blocks using the Woodbury matrix identity so the inverse is factorized into matrices $C_1$ and $C_2$ using a novel method called \textit{covariance factorization}. The size of $C_1$ and $C_2$ is equal to a mini-batch size, i.e. $m\times m$,  which is typically in the order of hundreds hence considerably smaller and easier to inverse compared to Kronecker factors in other NGD methods. Other matrices, i.e. $A'$ and $B'$, involved in the Fisher-block inverse in TENGraD, are \textit{reused}. This accelerates the inverse computation as it removes extra passes on the network via storing intermediate values computed during the forward-backward passes. As a result, TENGraD becomes an actual time-efficient approximate NGD method and improves the overall time.
 Also, to the best of our knowledge, we are the first to prove the linear rate convergence of approximate NGD methods for deep neural networks with ReLU activation functions. %We show that the number of iterations for TENGraD to converge to global optimum is $\mathcal{O}\left(\log(1/ \pngilon)\right)$ which outperforms other time-efficient approaches such as KFAC and KBFGS.  
 Our contributions are:
\begin{itemize}
    
    \item A time-efficient NGD method called TENGraD that  computes the exact inverse of Fisher blocks efficiently by using a \textit{covariance factorization} and \textit{reuse} technique. % along with   \textit{reuse} of intermediate values in the forward/backward pass of a neural network.
    Extension of TENGraD to convolutional layers  using tensor algebra. %Specifically, we show that our method is general and can be applied to a large class of linear transformations such as group convolution[] or layer normalization[].

    \item Empirical demonstration that TENGraD outperforms overall time of state-of-the-art NGD methods and SGD on well-known deep learning benchmarks.
    
    \item TENGraD converges to the global optimum with a linear rate if exact gradients are used with an iteration complexity of $\mathcal{O}\left(\log(1/ \epsilon)\right)$ which is comparable to exact NGD methods.
    
    %We also show that the maximal step size of approximate natural gradient descent for a network with $L$ layers is $\mathcal{O}\left(1/L)\right)$.
    
\end{itemize}

\vspace{-0.4cm}
\section{Background and Definitions}
%We begin by defining the basic notation for
%neural networks which we will use throughout this paper. 
 A deep neural network transforms its input $x$ to an output $f(\bm{W};\bm{x})$ through a series of $L$ layers where $\bm{W}$ is the vector of network parameters. During the forward pass,  layer $l$ performs a linear transformation on its input of $m$ samples, i.e.  $\mathcal{I}_{l} \in \mathbb{R}^{d_i^l \times m}$,  parameterized with $\bm{W}_{l} \in \mathbb{R}^{d_i^l \times d_o^l}$,  and computes its outputs or \textit{pre-activations}, $\mathcal{O}_{l} = \bm{W}_{l}^\top \mathcal{I}_{l}$, followed by a nonlinear activation function such as ReLU.  The optimal parameters are obtained by minimizing the average loss $\Lb$ over the training set during the backward pass:
\begin{equation}
{\small
    \Lb(\bm{W}) = \frac{1}{n}\sum_{i=1}^n \ell (f(\bm{W};\textbf{x}_i), \textbf{y}_i)
    \label{eq:loss_equation}
    }
\end{equation}
where $\bm{W} \in \mathbb{R}^p$ is the vector containing all of the network’s parameters concatenated together, i.e. $[\text{vec}(\bm{W}_{1})^\top \text{vec}(\bm{W}_{2})^\top ... \text{vec}(\bm{W}_{L})^\top]^\top$, $\text{vec}(M)$ is the operator which vectorizes matrix $M$ by stacking columns together, and  $\ell$ measures the accuracy of the prediction (e.g. cross entropy loss). % See \autoref{fig:main_fc} for the illustration of a one-layer network. 

% The optimal parameters for each layer is obtained during the backward pass via minimizing a loss function averaged over dataset $\{(x_i, y_i)\}_{i=1}^{n}$, i.e. $\min_{\thetab} \frac{1}{n}\sum_{i=1}^{n}l(f(\theta; x_i),y_i)$. During the backward, the gradient of loss w.r.t to the parameters
%  Stochastic gradient descent achieves this goal by iteratively updating the parameter according to $\bm{W} \leftarrow \bm{W} - \eta \bm{g}$ where $\eta$ is the learning rate and 
\textit{\textbf{Gradient and Jacobian in DNN}}.
Optimization methods that aim to minimize the average loss in \autoref{eq:loss_equation} work with gradient computations. The average gradient over mini-batch $m$ is defined as $\bm{g} = \frac{1}{m}\sum_{i=1}^{m}\nabla_{\bm{W}}\ell (f(\bm{W};\textbf{x}_i), \textbf{y}_i)$ which is  obtained using standard backpropagation. In the backpropagation, each layer $l$ receives the gradient w.r.t its output, i.e. the \textit{pre-activation derivatives} $\mathcal{G}_l\in \mathbb{R}^{d_o^l \times m}$, and computes the gradient w.r.t its parameters, i.e. $\bm{g}_l = \frac{1}{m} \mathcal{I}_l \mathcal{G}_l^\top$. 
%
% \textbf{Jacobian in DNN}. 
The Jacobian of loss w.r.t the parameters for a single output network is defined as $\bm{J} = [\bm{J}_1^{\top}, ..., \bm{J}_n^{\top}]^\top \in \mathbb{R}^{n \times p}$ where $\bm{J}_i$ is the gradient of loss w.r.t the parameters. In the stochastic setting, for a layer $l$ the Jacobian of loss w.r.t  the layer parameters is approximated via $\bm{J}_l = (\mathcal{I}_l \ast \mathcal{G}_l)^\top \in \mathbb{R}^{m \times d_i^l d_o^l}$. 

% .  \todo[inline]{This is the Jacobian of minibatch. When we use Jacobian we usually use if for all datasets. So it could be better to introduce something like $J^{B}$ to emphisize this is the batch.}

\textit{ \textbf{NGD update rule}}. Natural gradient descent scales the gradient with the inverse of FIM as follows:
% \vspace{-0.1cm}
\begin{equation}
{\small
    \bm{W}(k+1) = \bm{W}(k) - \eta \left(\bm{F}(\bm{W}(k))  + \alpha \bm{I}\right)^{-1} \bm{g}(k)
    \label{eq:update_rule_fisher}
    }
\end{equation} 
 where $\eta$ is the learning rate, $\alpha$ is the damping parameter, $\bm{{F}}$ is the FIM of the model's conditional distribution $P_{\textbf{y}|\textbf{x}}$ and is defined as $\bm{{F}} = \mathbb{E}_{ P_{(\textbf{x},\textbf{y})}}[\nabla_{\bm{W}} \log p_{\bm{W}} (\textbf{y}|\textbf{x}) \nabla_{\bm{W}} \log p_{\bm{W}} (\textbf{y}|\textbf{x})^\top]$  where $p_{\bm{W}}$ is the model's density function. Since $\bm{{F}}$ is singular for over-parameterized models, a non-negative damping term $\alpha$ is added to make it invertible. As shown in \cite{martens2020new}, the FIM coincides with Gauss-Newton matrix if the conditional distribution is in the exponential family which implies:
\begin{equation}
{\small 
    \bm{{F}} (\bm{W}(k)) = {\frac{1}{m}\sum_{i=1}^n \nabla_{\bm{W}} \ell_i {\nabla_{\bm{W}} \ell_i}^\top \big|_{
    \bm{W} = \bm{W}(k)}} = \frac{1}{n} \bm{J}(k)^\top \bm{J}(k) 
}
    \label{eq:fisher_matrix}
\end{equation}
where $\ell_i = \ell(f(\bm{W};\textbf{x}_i),\textbf{y}_i)$, $\bm{J}(k)=[\bm{J}_1(k)^\top,...,\bm{J}_n(k)^\top]^\top$ and $\bm{J}_i(k) = \nabla_{\bm{W}} \ell_i \big|_{
    \bm{W} = \bm{W}(k)}$. Hereafter, we use  $  \bm{{F}} (k) = \bm{{F}} (\bm{W}(k))$ to simplify the notations.

\textit{\textbf{Notations}}. Following notations are used in the paper:  $[L]:=\{1,...,L\}$, $\text{Diag}_{i=[L]}(A_i)$ is the block diagonal matrix with matrices $\{A_1,...,A_L\}$ on its diagonal; $\bm{X}\in\mathbb{R}^{n\times d}$ is the input data, i.e. $\bm{X}=[x_1,...,x_n]^\top$.  $\lambda_{\min}(M)$ is the smallest eigenvalue of  matrix $M$; Hadamard product between two matrices is  $\odot$;  the  Kronecker product is $\otimes$;  $\Vert.\Vert_2$ for the Euclidean norm of vectors/matrices; and $\ast$ is for the column-wise Khatri-Rao product; $A_{:i}$ is used to denote the $i$-th columns of matrix $A$.

\vspace{-0.2cm}
\section{Time-Efficient NGD with Exact Fisher-block Inversion}\label{Sec: TimeEfficientInversion}
 The computation bottleneck in NGD is $(\bm{F}(k)+\alpha I)^{-1}$ and its gradient product. Naively computing the Fisher inverse is impractical due to its large size, i.e. $p^2$ where $p$ is the number of parameters and can be millions. Therefore,  the Fisher matrix is typically approximated with a block-diagonal matrix $\bm{\hat{F}}=\text{Diag}\left( \bm{F}_1, ..., \bm{F}_L\right)$ where block $\bm{F}_{l}$ corresponds to layer $l$ \cite{grosse2016kronecker}. As a result, $(\bm{\hat{F}}+ \alpha I)^{-1}$ becomes a block diagonal matrix.  TENGraD computes the exact inverse of each block, i.e. $(\bm{F}_{l}+ \alpha I)^{-1}$ (hereafter, all computations are performed on one block, thus,  $l$  is removed from notations). TENGraD uses the Woodbury identity for each Fisher-block:

% Therefore,  the Fisher matrix is typically approximated \todo{Can add reference here?} with a block-diagonal matrix \bc{$\hat{\bm{F}}(k):=\hat{\bm{J}}(k)^\top \hat{\bm{J}}(k)$ where $\hat{\bm{J}}(k)=\text{Diag}_{l\in[L]}\{J_{l}(k)\}$ for $J_{l}(k):=(\frac{\partial \bm{f}(\bm{W}(k))}{\partial \text{vec}(W_l)})$ and $\bm{f}(\bm{W})=[f(\bm{W};x_1),..,f(\bm{W};x_n)]^\top$.} As a result, $(\hat{\bm{F}} + \alpha I)^{-1}$ becomes a block diagonal matrix and TENGraD computes the exact inverse of each block, i.e. $(J_{l}^\top(k) J_l(k)+ \alpha I)^{-1}$. \bc{Hereafter, all computations are performed on one block at each iteration, thus the layer index $l$ and iteration $k$ are removed to simplify notations whenever it is implied from the context.}

% +  These Fisher-blocks and their inverses are still computationally expensive as their dimension scales with layer's size. While many work aim to approximate the block inverses, TENGraD leverages the Fisher block structures and computes the exact Fisher block inverses by applying the Woodbury matrix identity on the NGD update rule for each block. 

% The direct advantage of using this identity is that the higher order term in the computation complexity of direct block-inversion is cubic in the parameter dimensions, i.e. $\mathcal{O}(p^3)$,  which after applying Woodbury becomes cubic in the batch size dimension, i.e. $\mathcal{O}(m^3)$.Therefore, as long as the batch size $m$ is smaller compared to parameters dimension, which typically holds for overparametrized models, it can lead to a better performance.

\begin{equation}
{\footnotesize
\label{eq:smw}
    \bm{W}(k+1) = \bm{W}(k) - \frac{\eta}{\alpha} \left(\bm{g}(k) - \underbrace{\frac{\bm{J}(k)}{m}^\top}_{C} \underbrace{\left(\frac{\bm{J}(k) \bm{J}(k)^\top}{m} +\alpha I\right)^{-1}}_{A} \underbrace{\bm{J}(k) \bm{g}(k)}_{B}\right) 
}
\end{equation}
 where $\bm{W}, \bm{g}$, and $\bm{J}$ are the parameters, gradient, and Jacobian associated with the Fisher-block (we drop the index $k$ to simplify the notations).
Main steps of the new update rule are:
\textit{(A)} computing and inverting the \textit{Gram Jacobian}
 $\bm{J}\bm{J}^\top$, which is costly because the Jacobian  $\bm{J}$ and a matrix-matrix multiplication have to be computed, \textit{(B)} scaling  of the gradient  $\bm{g}(k)$ using the Jacobian to transform the gradient into input space, and \textit{(C)} rescaling with $\bm{J}^\top$ to transform back to the parameter space. The Gram matrix $\bm{J}\bm{J}^\top$ computation in \textit{(A)} has high complexity and is proportional to both input and output dimensions of a layer, i.e. $\mathcal{O}(m d_i d_o)$.  TENGraD mitigates this cost with \textit{covariance factorization} that factors the Gram Jacobian into a low-cost  operation between two small matrices. Steps \textit{(B)} and \textit{(C)} also involve Jacobian vector products that are orders of magnitude more computationally expensive than a gradient computation \cite{dangel2019backpack}. TENGraD mitigates this cost by \textit{reusing} the input and activations computed during the forward-backward pass. %Section 3.1 discusses these approaches in TENGraD for a fully connected layer along with complexity analysis, we then provide an extension to convolutional layers in Section 3.2.  

\subsection{Covariance Factorization and Reusing of Intermediate Values in TENGraD}

%TENGraD covariance factorization and reuse method  improve computational efficiency in  \autoref{eq:smw}. % of intermediate variables method that reduces the computational and storage complexity in \autoref{eq:smw}.

%In this section, we first show how covariance factorization helps to reduce the computation complexity of Gram Jacobian $JJ^\top$ by leveraging the Fisher block structures, in particular its low-rank property. Then we show that reusing intermediate variables can further reduce the computation cost by breaking down the Jacobian vector products into two smaller matrix multiplications.

%\textit{Covariance Factorization}. 
TENGraD factors the Gram matrix $\bm{J}\bm{J}^\top$ into two smaller $m\times m$ matrices $C_1$ and $C_2$ called covariance factors such that $\bm{J}\bm{J}^\top = C_1 \odot C_2$, $m$ is the batch size. These factors can be obtained with low overhead as batch size is typically small. % very small and also they do not require the explicit computation of Jacobian matrix.
%
%
%
%
%
% The covariance factorization leverages the structure of gradients of a layer, in particular its low-rank property, and  mitigates the cost of computing the Gram Jacobian by first avoiding the explicit computation of $J$ and second factoring the Gram matrix into low-cost element wise operation of its smaller factors.
%
 The basic method for computing the factors was first introduced in \cite{ren2019efficient} for fully connected networks which we also add here for the sake of completeness. For only one sample and for the fully connected layer in \autoref{fig:main_fc}, with the input  $\mathcal{I}$ and the pre-activations  $\mathcal{O} = \bm{W}^\top \mathcal{I}$, we show the low-rank property of the gradient and how it is leveraged to reduce the computation complexity of the Gram Jacobian. The gradient of the loss w.r.t parameter $\thetab_l$ for input sample $i$ is computed using the input and pre-activation derivatives by $ \bm{g}_i = \mathcal{I}_{:i} \mathcal{G}_{:i}^\top$.  The gradient $\textbf{g}_i$ is the outer product of two vectors and therefore a rank-one matrix. The $(i,j)$ element of Gram Jacobian is the inner product of vectorized gradients of two samples $i$ and $j$, i.e. $[\bm{J}\bm{J}^\top]_{(i,j)} = \langle \text{vec}(\bm{g}_i), \text{vec}(\bm{g}_j) \rangle$. With  $\text{vec}(uv^\top)= v \otimes u$,  the Gram Jacobian becomes $[\bm{J}\bm{J}^\top]_{i,j} = \langle \mathcal{I}_{:i} \otimes \mathcal{G}_{:i} , \mathcal{I}_{:j} \otimes \mathcal{G}_{:j} \rangle$ and with $\langle u_1 \otimes v_1 , u_2 \otimes v_2 \rangle  =  u_1^\top u_2 \cdot v_1^\top v_2$, it is rewritten as  $[\bm{J}\bm{J}^\top]_{i,j} = \mathcal{I}_{:i}^\top \mathcal{I}_{:j} \cdot \mathcal{G}_{:i}^\top \mathcal{G}_{:j}$. As a result, the $(i,j)$-th element of the Gram Jacobian is efficiently computed without forming the gradients.  Extended to a mini-batch of $m$ samples, the per-sample gradients are written as the column-wise Khatri-Rao product of input and pre-activation derivatives, i.e.  $\bm{J} = \left(\mathcal{I} \ast \mathcal{G}\right)^\top$. Therefore the compact form of the Gram Jacobian is:
% \begin{equation}
% \label{eq:fc_cfac}
%     J = \left(\mathcal{I} \sbullet[1.25] \mathcal{G}\right) \Rightarrow
%     JJ^\top = \left(\mathcal{I} \sbullet[1.25] \mathcal{G}\right) \left(\mathcal{I}^\top \ast \mathcal{G}^\top \right) = \underbracket[0.127ex]{\mathcal{I} \mathcal{I}^\top}_{\Sigma_{I}} \odot \underbracket[0.127ex]{\mathcal{G} \mathcal{G}^\top}_{\Sigma_{O}}
% \end{equation}
\begin{equation}
\label{eq:fc_cfac}
{ \small 
    \bm{J} = \left(\mathcal{I} \ast \mathcal{G}\right)^\top \Rightarrow
    \bm{J}\bm{J}^\top = \left(\mathcal{I} \ast \mathcal{G}\right)^\top  \left(\mathcal{I} \ast \mathcal{G}\right) = \underbracket[0.127ex]{\mathcal{I}^\top \mathcal{I}}_{C_1} \odot \underbracket[0.127ex]{\mathcal{G}^\top \mathcal{G}}_{C_2}
}
\end{equation}
 From \autoref{eq:fc_cfac}, the Gram Jacobian is written as the Hadamard product of two smaller $m\times m$ matrices, i.e. covariance factors. The input covariance $C_1$ and the pre-activation covariance  $C_2$ are efficiently computed during the forward and backward pass of the neural network using a matrix-matrix multiplication. The Jacobian does not need to be explicitly formed to compute these factors,   reducing the overall computation complexity. Also, these factors are smaller and have low storage costs compared to Kronecker-factors, which have a size of $d_i \times d_i$ and $d_o \times d_o$.

%Steps \textbf{B} and \textbf{C} during the NGD update iterations in \autoref{eq:smw} is considerable, in particular $m$ times the number of parameters. 

%\textbf{Efficient Reuse of Intermediate Values}. 
%To minimize the overheads of  NGD, TENGraD   applies the  NGD  update every $T$ iterations  of DNN training. 
%NGD updates require the Jacobian to  compute steps \textit{(B)} and \textit{(C)} of \autoref{eq:smw}.
The cost of storing the Jacobian for the NGD update in steps \textit{(B)} and \textit{(C)} of \autoref{eq:smw} is considerable because its storage cost is $m$ times the number of parameters. 
With a reuse method, TENGraD instead recomputes the Jacobian when needed, with a negligible computation cost and by storing small data structures.   TENGraD breaks down the computation of Jacobian vector products in steps \textit{(B)} and \textit{(C)} into two operations involving input and pre-activation derivative  matrices  $\mathcal{I}$ and $\mathcal{G}$.
%
%Applying \autoref{eq:smw} and in general any NGD update rule is an expensive operation that needs to be amortized. Therefore, the NGD update rule is applied every $T$ iterations which is common among all approximate NGD methods and TENGraD also adopts the same strategy. However, the cost of storing Jacobian to perform steps \textbf{B} and \textbf{C} in the update rule in \autoref{eq:smw} is considerable, in particular $m$ times bugger than the size of layer parameters. Hence, 
%With a reuse method, TENGraD adopts a reuse method that enables recomputing the Jacobian matrix with a low storage cost with a negligible computation cost. In order to apply the reuse method, TENGraD breaks down the computation of Jacobian vector products in steps \textbf{B} and \textbf{C} into two operations involving input and pre-activation derivative  matrices  $\mathcal{I}$ and $\mathcal{G}$.
%
% particular performing Jacobian vector products is expensive if one explicitly computes the Jacobian $J$ as it is $\mathcal{O}(m)$ times more expensive than computing a gradient. A common approach to avoid explicit formation of $J$, is using auto differentiation frameworks to perform backpropagation which can lead to significant overheads due to recomputing pre-activation derivatives and extra passes on the network[].
%
% TENGraD aims to reduce the complexity of these computation by storing and reusing intermediate values computed  during forward and backward pass of the network.
%
In step \textit{(B)} in \autoref{eq:smw}   the vectorized form of the gradient for the layer parameters, i.e. $v = \text{vec}(\bm{g}(k))$, is propagated. The objective is to compute $\bm{J} \text{vec}(\bm{g}(k))$ without explicitly forming the Jacobian. With  $\bm{J} = \left(\mathcal{I} \ast \mathcal{G}\right)^\top$ and using the properties of column-wise Khatri-Rao product:
\vspace{-0.05cm}
\begin{equation}
{\small
    \bm{J} \text{vec}(\bm{g}(k)) = \left(\mathcal{I} \ast \mathcal{G}\right)^\top \text{vec}(\bm{g}(k))= \left(\left( \bm{g}(k)^\top \mathcal{I} \right) \odot \mathcal{G} \right) \mathbf{1},
    \label{eq:jvp_fc}
    }
\end{equation}
\vspace{-0.05cm}
where $\mathbf{1}$ is a vector of all ones with appropriate dimension. The Jacobian vector product can be computed with two efficient matrix operations without forming the Jacobian. First,  $v_1$ is computed with $v_1 =  \bm{g}(k)^\top \mathcal{I}$, followed by a Hadamard product with the pre-activation derivative matrix, i.e.  $v_2 = v_1 \odot \mathcal{G}$, and a column-wise summation on $v_2$. 
%
% As shown, TENGraD uses layer dimensions to choose between the  Khatri-rao product varaints in  \autoref{eq:jvp_fc} which further reduces the computation complexity, chooses between each implementation based on the layer dimensions to further reduce the computation complexity. If the layer acts as a bottleneck,i.e. input dimension is larger than output dimension, the first implementation is picked. 
%
%
%
% Considering the first implementation in \autoref{eq:jvp_fc},
%
%
%
In \textit{(C)} in \eqref{eq:smw}, the objective is to compute $\bm{J}^\top v$ with $v\in \mathbb{R}^m$ obtained from step \textit{(A)}. Using the column-wise Khatri-Rao structure of $\bm{J}$: 
\vspace{-0.05cm}
\begin{equation}
{\footnotesize
    \bm{J}^\top v =   \left(\mathcal{I} \ast \mathcal{G}\right)v  =  \mathcal{I} \left(  v \mathbf{1}^\top \odot \mathcal{G}^\top\right).
    \label{eq:vjp_fc}
}
\end{equation}
\vspace{-0.05cm}
Similarly,  $\bm{J}^\top v$ is computed  using the two step process in \autoref{eq:vjp_fc}. By storing  $I$ and  $G$, TENGraD can efficiently compute the required operations in the NGD update. 

\textbf{Storage and Computation Complexity}.
\autoref{tab:complexity} shows the computation and storage complexity of TENGraD vs other NGD methods.  \textit{Curvature} shows the cost of computing and inverting Kronecker-factors for KFAC, EKFAC, and KBFGS, and Gram Jacobian for TENGraD. The computation complexity of covariance factors is $\mathcal{O}(m^2 ( d_i + d_o))$, while Kronecker factors in KFAC have a complexity of $\mathcal{O}(m d_i^2 + m d_o^2)$. Inverting the factors in TENGraD is $\mathcal{O}(m^3)$ while  that of KFAC is $\mathcal{O}(d_i^3 + d_o^3)$. Thus, when the batch size is smaller than layer dimensions TENGraD's computation complexity is better than others.  TENGraD reduces storage complexity with reuse from  $\mathcal{O}(m d_i d_o)$ to $\mathcal{O}(m (d_i +  d_o))$, see the \textit{Extra Storage} column.  \textit{Extra Pass} refers to the cost of  a second backpropagation and \textit{Ste}p is the computation cost of parameter updates after computing the curvature.
%As shown in this section, TENGraD significantly can reduce the computation complexity of NGD computations. By using the covariance factorization method, the computation complexity of covariance factors significantly reduces from $\mathcal{O}(m^2 d_i d_o)$ to $\mathcal{O}(m^2 ( d_i + d_o))$. A similar analysis shows that computing Kronecker factors in KFAC has a complexity of $\mathcal{O}(m d_i^2 + m d_o^2)$. Moreover, inverting the factors in TENGraD has a complexity of $\mathcal{O}(m^3)$ compared to KFAC which has a complexity of $\mathcal{O}(d_i^3 + d_o^3)$. Therefore TENGraD shows a better computation complexity as long as the batch size is smaller compared to layer dimensions. Moreover, TENGraD reduces the computation complexity by applying the reuse of intermediate values method from  $\mathcal{O}(m d_i d_o)$ to $\mathcal{O}(m (d_i +  d_o))$ which is considerably lower compared to all Kronecker-factored methods. 
 %The summary of our complexity analysis is shown in \autoref{tab:complexity} . In \autoref{tab:complexity}, the "Extra Pass" refers to the cost of performing a second backpropagation to estimate the second order information. The "Curvature" refers to the cost of computing and inverting Kronecker-factors for KFAC, EKFAC and KBFGS and Gram Jacobian for TENGraD. The step also refers to the computation cost for the updating the parameters after computing the curvature. 
% \vspace{-0.5cm}
\renewcommand{\tabcolsep}{2pt}
\begin{table}[h!]
\centering
\caption{Computation and storage complexity of NGD methods.}
\label{tab:complexity}
\begin{tabular}{|l|c|c|c|c|}
\hline
Algorithm & Curvature                               & Extra Pass               & Step                                     & Extra Storage                       \\ \hline
TENGraD       & $\mathcal{O}(m^2 d_i+m^2 d_o+m^3)$      & $\mathcal{O}(m d_i d_o)$ & $\mathcal{O}(m d_i d_o + m d_o + m^2)$   & $\mathcal{O}(m d_i + m d_o + m^2 )$ \\
KFAC &
  $\mathcal{O}(m d_{i}^2+m d_{o}^2+ d_{i}^3 + d_{o}^3)$ &
  $\mathcal{O}(m d_i d_o)$ &
  $\mathcal{O}(d_{i}^2 d_o + d_{o}^2 d_i)$ &
  $\mathcal{O}(d_i^2 + d_o^2 )$ \\
EKFAC &
  $\mathcal{O}(m d_{i}^2+m d_{o}^2+ d_{i}^3 + d_{o}^3)$ &
  $\mathcal{O}(m d_i d_o)$ &
  $\mathcal{O}(m d_o d_{i}^2 + m  d_{o}^2  d_i)$ &
  $\mathcal{O}(m d_i d_o + d_i^2 + d_o^2 )$ \\
KBFGS     & $\mathcal{O}(m d_i^2 + m d_o + d_{o}^2)$ & $\mathcal{O}(m d_i d_o)$ & $\mathcal{O}(d_{i}^2 d_o + d_{o}^2 d_i)$ & $\mathcal{O}(d_i^2 + d_o^2 )$       %\\
%SGD       & ---                                     & ---                      & $\mathcal{O}(d_i d_o)$                   & ---                               
\\ \hline
\end{tabular}
\end{table}
%  As seen in \autoref{tab:complexity}, computing the curvature in TENGraD has a linear dependency to input and output dimension while KFAC and EKFAC has a cubic dependence on $d_i$ and $d_o$. Moreover, curvature computation in KBFGS has a quadratic dependence on $d_i$ and $d_o$. Similarly, TENGraD has less computation per iteration  for updating parameters, i.e. "Step", while all other methods except SGD has a quadratic dependence on input and out dimensions. Moreover, the storage requirement for TENGraD are much lower compared to other methods for moderate batch sizes. Overall, TENGraD outperforms all other second order methods in computation complexity. 
\begin{figure}
     \centering
        \includegraphics[width=0.7\textwidth]{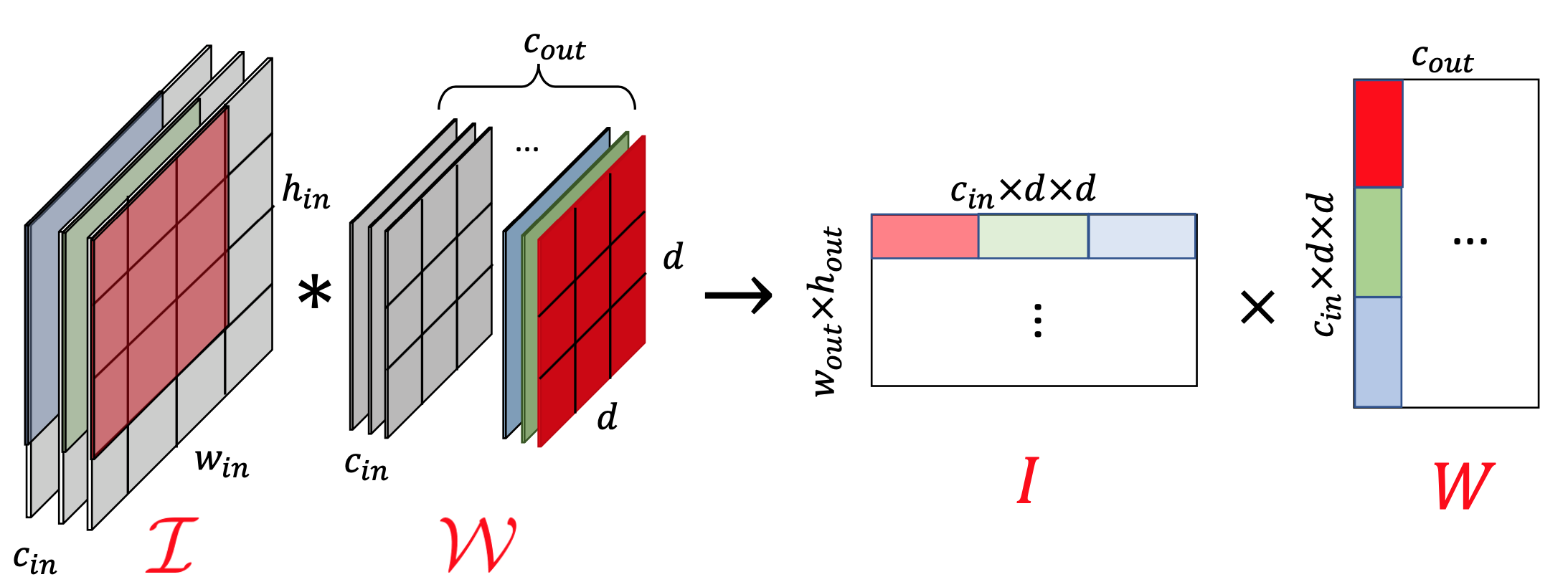}
         \caption{Convolution shown as matrix-matrix product of unfolded input and weights.}
     \label{fig:conv_figure}
     \setlength{\belowcaptionskip}{-4pt}
\end{figure}

% \setstretch{2}
\vspace{-0.3cm}
\subsection{Extension to Convolution Layers}
We extend TENGraD to support convolutional layers, general form provided in the Appendix. %With a brief background on convolutional layers, first we demonstrate that convolution operation can be written as a matrix multiplication which will be used to simplify the derivation of Jacobian matrix. Afterward, we derive the covariance factorization for the convolution layer by extending to tensors and show that the the Gram Jacobian can be represented by two smaller tensors. % While fully connected layers play an important role in understanding neural networks, they have limited practicality compared to convolution layers[]. Therefore, it is crucial to derive efficient implementation of NGD steps in \autoref{eq:smw} for convolutional layers. However, this task is not trivial as 
Convolutional layers operate on high-dimension tensors. For a general tensor $\mathcal{A} \in  \mathbb{R}^{N_1 \times ...\times N_p}$  with dimension $p$, the components of a tensor  can be represented with $\mathcal{A}_{i_1, ...,i_p}$ where index $i_\alpha$ is associated with dimension $\alpha$. We consider a convolutional layer with the input tensor $\mathcal{I} \in \mathbb{R}^{c_{in}\times w_{in} \times h_{in}}$ and convolutional filters 
$\mathcal{W} \in \mathbb{R}^{c_{in} \times d \times d \times c_{out}}$. For the weight tensor $\mathcal{W}$ the spatial support of each kernel filter is $d\times d$. There are $c_{in}$ input channels and $c_{out}$ feature maps. Let 
$\mathcal{O}\in \mathbb{R}^{c_{out}\times w_{out}\times h_{out}}$ be the output of the convolutional layer, where $w_{out} = \lfloor w_{in}+2p_d-d \rfloor /s + 1$ and $h_{out} = \lfloor h_{in}+2p_d-d\rfloor/s + 1$. Also, $p_d$ and $s$ are the padding and stride parameters, respectively.

To extend covariance factorization for  $\bm{J}\bm{J}^\top$ in a convolutional layer, we first  derive a closed form equation for the Jacobian and show that the convolution operation can be written as a matrix multiplication for a single sample. 
From \autoref{fig:conv_figure}, $\mathcal{I}$ can be reshaped to a matrix $I$ with dimensions $(w_{out} \times h_{out}) \times (c_{in} \times d \times d)$, each row is a unfolded  sub-tensor in $I$ that has the same size of a group
of $c_{in}$ filters. Each $c_{in}$ filter is also unfolded into a 1-D vector with size $(c_{in} \times d \times d) \times 1$. The weight tensor $\mathcal{W}$ is reshaped to a matrix $W$ with size $(c_{in} \times d \times d) \times c_{out}$  (see \autoref{fig:conv_figure}). Hence, convolution  can be written as $O = I \cdot W$,  where the shape of $O$ is $(w_{out} \times h_{out}) \times (c_{out})$. 
Matrix $O$ is reshaped to the output tensor $\mathcal{O}\in \mathbb{R}^{c_{out}\times w_{out}\times h_{out}}$ by folding each column into a $w_{out}\times h_{out}$ feature map. Applying a similar unfolding to the pre-activation derivatives tensor  $\mathcal{G}\in \mathbb{R}^{c_{out}\times w_{out}\times h_{out}}$  results in the matrix $G$ with a shape of $(w_{out} \times h_{out}) \times (c_{out})$.

%With the above matrix form of convolution, 
The compact form of the Jacobian  for a batch of $m$ samples, with $O = I \cdot W$, 
%which is the matrix form of gradients can be written 
is:
% the gradient of loss w.r.t the filters  can be written as $\bm{g} = I^\top \cdot G$.
\begin{equation}
{\small
   \bm{J} = [\text{vec}(\bm{g}_1), ..., \text{vec}(\bm{g}_m),] = [\text{vec}(I_1^\top \cdot G_1),..., \text{vec}(I_m^\top \cdot G_m)]^\top
   \label{eq:tensor_jacobian}
  }
\end{equation}
 where $I_i$ and $G_i$ are the unfolded input and output for sample $i$. We derive a closed form for an arbitrary element $(i,j)$ in the Gram Jacobian matrix and then extend to the whole batch size. Using \autoref{eq:tensor_jacobian} each element in the Gram Jacobian is written as:
\begin{equation}
    [\bm{J}\bm{J}^\top]_{i,j} = \langle \text{vec}(I_i^\top \cdot G_i), \text{vec}(I_j^\top \cdot G_j) \rangle = \bm{1}^\top I_i I_j^\top \cdot G_i G_j^\top \bm{1}
    \label{eq:conv_jacobian}
\end{equation}
 Hence, each element $(i,j)$ in Gram Jacobian is obtained via a reduction on covariance matrices $I_i I_j^\top \in \mathbb{R}^{(w_{out}\times h_{out}) \times (w_{out}\times h_{out}) }$ and $G_i G_j^\top \in \mathbb{R}^{(w_{out}\times h_{out}) \times (w_{out}\times h_{out})}$. The covariance matrices are defined over the spatial support $S = w_{out}\times h_{out}$.  

With tensor notations, we expand the dimensions of covariance factors. Since  element $(i,j)$ in  Gram Jacobian is represented with two covariance matrices, the $m \times m$ Gram Jacobian is   $m^2$ covariance matrices. The covariance tensors $C_1 \in \mathbb{R}^{m\times m \times S  \times S}$ and $C_2 \in \mathbb{R}^{m\times m \times S  \times S}$ are defined as:
\begin{equation}
{\small
    [C_1]_{i,j,*,*} = I_i I_j^\top, \quad [C_2]_{i,j,*,*} = G_i G_j^\top
    }
\end{equation}
Where $[C_1]_{i,j,*,*}$ denotes the sub-tensor at  $(i,j)$. As a result the Gram Jacobian is written as:
\begin{equation}
{\small
    \bm{J}\bm{J}^\top = \sum_{s \in S} \sum_{s' \in S} [C_1 \odot C_2]_{*,*,s,s'}
    \label{eq:tensor_facotrization}
}
\end{equation}

%Therefore the Gram Jacobian matrix is the Hadamard product of two smaller tensors followed by a reduction on spatial dimensions. Interestingly, this result, which is similar to the one for fully connected layers, holds for any linear transformation (see Appendix). In the following corollary we show that covariance factorization reduces the computation complexity for convolutional layers with small spatial supports. 

\begin{corollary}
For a standard convolutional layer with spatial support $S = w_{out} \times h_{out}$, overall filter support $F = c_{in}\times d\times d$ and batch size $m$, TENGraD reduces the computational complexity of the Gram Jacobian computation from $\mathcal{O}(m^2 F c_{out} + m F S c_{out})$ to $\mathcal{O}(m^2 S^2 (F + c_{out}))$.
\end{corollary}

%\textbf{Efficient Reuse of Intermediate Values}:

For TENGraD's reuse for step \textit{(B)}, the gradient tensor of the convolution layer, i.e.  $\bm{g}(k) \in \mathbb{R}^{c_{in} \times d \times d  \times c_{out}} $ is reshaped to $\bm{\hat{g}}(k) \in \mathbb{R}^{(c_{in} \times d \times d) \times  c_{out} }$ to compute $\bm{J}v = \bm{J} \text{vec}(\bm{\hat{g}}(k))$.  Using \autoref{eq:conv_jacobian}, this can be written as a two step process without explicitly forming the Jacobian matrix: $ [\bm{J}v]_i = \bm{1}^\top \left( G_i \cdot \bm{\hat{g}}(k)^\top  \odot I_i \right)  \bm{1} $.  Similarly, for step \textit{(C)},  $\bm{J}^\top v$ for $v\in \mathbb{R}^m$ is computed with $\bm{J}^\top v = \sum_{i=1}^m I_i^\top (G_i \cdot v_i)$. For better storage, instead of $I$ we store the input tensor $\mathcal{I}$. % and apply the unfolding before it is used. %Notice that unfolding is a cheap operation and has a very low overhead. 

\vspace{-0.2cm}
\section{Linear Convergence in TENGraD} We provide convergence guarantees for TENGraD with exact gradients (i.e. full batch case with $m=n$). We focus on the single output but a general case with multiple outputs will be similar. As shown TENGraD, converges linearly at a rate independent from the condition number of the input data matrix $\bm{X}^\top \bm{X}$, unlike the convergence rate provided for KFAC at \cite{NEURIPS2019_1da546f2}. Consequently, for sufficiently ill-conditioned data, our convergence guarantees will improve upon those currently available for KFAC.\looseness=-1

Lets us introduce the output vector  $\bo{u}(\bm{W}) = [\normalfont{u}_1(\bm{W}), ..., \normalfont{u}_n (\bm{W}) ]^\top$ where $\normalfont{u}_i(\bm{W}) = f_L(\bm{W}; x_i)$ and $y = [\normalfont{y}_1, ..., \normalfont{y}_n ]^\top$. We consider the squared error loss  $\mathcal{L}$ on a given dataset $\{x_i,y_i \}_{i=1}^{n}$ with $x_i \in\mathbb{R}^{d}$ and $y_i \in \mathbb{R}$, i.e. the objective is to minimize
$
\min_{\bm{W} \in \mathbb{R}^{p}} \mathcal{L}(\bm{W})= \frac{1}{2}\Vert \bo{u}(\bm{W})-y\Vert^2
$. The update rule \eqref{eq:smw} of TENGraD with exact gradient becomes
\vspace{-0.2cm}
\begin{equation}
{\small
    \bm{W}(k+1) = \bm{W}(k) - \eta \left(\bm{\hat{F}}(\bm{W}(k))  + \alpha \bm{I}\right)^{-1} \bm{J}(k)^\top (\bo{u}(\bm{W}(k))-y),
    \label{eq:apx_update_rule}
}
\end{equation}
where $\hat{\bm{F}}(\bm{W}(k)):=\hat{\bm{J}}(k)^\top \hat{\bm{J}}(k)$ is the Fisher-block matrix and the block Jacobian is defined as  $\hat{\bm{J}}(k)=\text{Diag}_{l\in[L]}\left( J_{l}(k)\right)$ for $J_{l}(k):=(\frac{\partial \bm{f}(\bm{W}(k))}{\partial \text{vec}(W_l)})$ and $\bm{f}(\bm{W})=[f_L(\bm{W};x_1),..,f_L(\bm{W};x_n)]^\top$. It can be seen directly from the TENGraD update rule \eqref{eq:smw} that if the Gram matrix $G_l(k):= J_{l}(k)J_l(k)^\top$ stays positive definite and bounded for each layer $l$ and iteration $k$, then by the standard theory of preconditioned gradient descent methods \cite{bertsekas1997nonlinear}, it will converge globally for sufficiently small stepsizes. In the following, we introduce two assumptions; the first one ensures that at the initialization Gram matrix is positive-definite and the second assumption is a stability condition on the Jacobian requiring that it does not vary too rapidly. These assumptions will allow us to control the convergence rate.
\begin{assumption}\label{assump 1} The data $X$ is normalized, $|y_i|=\mathcal{O}(1)$, and for any $i\neq j$, $x_i$ and $x_j$ are independent. Lastly, the Gram matrix for individual layers are positive definite at initialization, i.e.  $\min_{l\in[L]}\lambda_{min}(G_l(0)) =  \lambda_{0} > 0$.
\end{assumption} 
\vspace{-0.2cm}
Lee et al. \cite{lee2019} shows that if the input data are i.i.d. samples, then Assumption \ref{assump 1} can often be satisfied.
\begin{assumption}\label{assump 2}
There exists $0<C\leq \frac{1}{2}$  that satisfies $\Vert{\bm{J}(\bm{W}(k))} - \bm{J}(\bm{W}(0))\Vert_2 \leq \frac{C}{3}\lambda_{0} ^ {\frac{1}{2}}$ if $\Vert \bm{W}(k) - \bm{W}(0)\Vert_2 \leq 3\lambda_{0} ^ {\frac{-1}{2}} \Vert \bm{y}-\bm{u}(0)\Vert_2 $.
\end{assumption} 
Notice that Assumption \ref{assump 2} requires that the network behaves like a linearized network for small values of constant $C$ \cite{lee2019}. As a result of Assumptions \ref{assump 1}-\ref{assump 2}, the Fisher block matrix remains close to the initialization over iterations and therefore the Gram matrices of layers stay positive-definite during training which is parallel to the setting of \cite{NEURIPS2019_1da546f2,du2018gradient,nguyen2020tight}. In practice, we could also keep Gram matrices well-conditioned during training by tuning the damping parameter. 

Next, we present our convergence result in Theorem \ref{thm: Convergence}. Our analysis follows the techniques of \cite{NEURIPS2019_1da546f2} and adapts them for TENGraD. Notice that TENGraD differs from NGD and KFAC in terms of the matrix it uses to approximate FIM; therefore results of \cite{NEURIPS2019_1da546f2} do not directly apply to our setting. As a proof technique, we first utilize the Assumptions \ref{assump 1} and \ref{assump 2} to derive a lower bound on the smallest eigenvalue $\lambda_0$ of the Fisher-block matrix $\hat{\bm{F}}$ showing that it is positive, and building on this result we derive the rate in the following result. The proof can be found in the Appendix.
\begin{thm}\label{thm: Convergence} \label{theorem1}
Suppose Assumptions \ref{assump 1}, \ref{assump 2} hold. Consider the update rule (\ref{eq:apx_update_rule}) for a network with $L$ layers, damping parameter $0 < \alpha \leq \frac{4\lambda_0}{9n}$, and learning rate  $\eta \in (0, \frac{2L\gamma -1 - 2C\sqrt{L}\gamma }{(L +C\sqrt{L}\gamma)^2 })$ where $\gamma = \frac{\lambda_0}{\lambda_0 + \frac{9}{4}n\alpha}$. Then, we have
%\vspace{-0.2c}
%\begin{equation*}
$  {\footnotesize
    \Vert \bo{u}(k) - \bo{y}\Vert_2^2 \leq (1 - \eta)^k\Vert \bo{u}(0) - \bo{y}\Vert_2^2.
    }$
% \end{equation*}
 \end{thm}
 
%\begin{proof} 
%See the Appendix.
%\end{proof}\todo{do you need $\eta>0$?}

\autoref{theorem1} states that TENGraD converges to the global optimum with a linear rate under Assumptions 1 and 2. Therefore, TENGraD requires only $\mathcal{O}(\log(1/\epsilon))$ iterations to achieve $\epsilon$ accuracy. Moreover, our result shows that a smaller learning rate is needed for a deeper network to guarantee linear convergence. This behavior is different than that of exact NGD because the rate of exact NGD provided in \cite{NEURIPS2019_1da546f2} does not depend on the network size but rather on the parameter $C$. We also observe from Theorem \ref{thm: Convergence} that smaller damping parameter improves the upper bound on the learning rate which suggests faster convergence for TENGraD. This makes the analysis of $\lambda_0$ important for the performance of TENGraD since Gram matrix can potentially be ill-conditioned. The work \cite{NEURIPS2019_1da546f2} uses matrix concentration inequalities and harmonic analysis techniques to bound the minimum eigenvalue of the Gram matrix of the L-layer ReLU network which does not provide a tight bound for the Gram matrix of each layer. The authors of \cite{nguyen2020tight} have obtained tighter bounds in the NTK setting,
and their result can be used to derive lower bound on $\lambda_{0}$ of $\hat{\bm{J}}_l \hat{\bm{J}}_l(k)^\top$. Due to limited space, we present the details on the explicit bound on $\lambda_0$ at Appendix. 

We can show that under some mild conditions on the distribution of input data $X$, if the weights of  layer $l$ are drawn from $\mathcal{N}(0,\beta_l^2)$, and the conditions $d_l=\tilde{\Omega}\left( N\right)$ and $\prod_{j=1}^{l-2}\log(d_j)=o(\min_{j\in[0,l-1]}d_j)
$ are satisfied,
then the smallest eigenvalue of Gram matrix $\hat{\bm{J}}_l(k)\hat{\bm{J}}_l(k)^\top$ satisfies $\lambda_{\min}(\hat{\bm{J}}_l(k)\hat{\bm{J}}_l(k)^\top)\geq C_{\beta}\Omega(\prod_{l=1}^{L}d_l)$ for all $k>0$  with some probability where the coefficient $C_\beta$ depends on the variances $\beta_{l}^2$ (see Appendix). This implies that we can choose a smaller damping parameter $\alpha$  to get a faster rate according to Theorem \ref{thm: Convergence} which, in turn, improves the performance of TENGraD. Otherwise,  $\alpha$ needs to be larger to guarantee the inversion at \eqref{eq:apx_update_rule}. 

\vspace{-0.3cm}
\section{Experimental Results}
\vspace{-0.2cm}
\label{sec:experiments}
We test TENGraD, KFAC, EKFAC, KBFGS, and SGD with momentum on three image classification benchmarks namely \verb+Fashion-MNIST+ \cite{xiao2017fashion}, \verb+CIFAR-10+\cite{Krizhevsky09learningmultiple} and \verb+CIFAR-100+) \cite{Krizhevsky09learningmultiple}. For \verb+Fashion-MNIST+ a network with its convolution layers and one fully connected layer is used, hence called \verb+3C1F+. For other benchmarks, the following DNN architectures are used \verb+DenseNet+ \cite{huang2017densely}, \verb+WideResNet+ \cite{zagoruyko2016wide} and \verb+MobileNetV2+ \cite{sandler2018mobilenetv2}. 
In all experiments, 50K samples are chosen for training and 10K for test. Experiments are conducted on AWS P3.2xlarge machines with one Volta V100 GPU and 16GB RAM. Each experiment is repeated 5 times and the average is reported. In our theoretical analysis, we assumed deterministic gradients but in the experiments we use stochastic gradients estimated with mini-batch size $m=128$. The Appendix  provides additional setup information.\looseness=-1

\begin{figure*}[h!]
    \centering
    \begin{subfigure}[b]{0.32\textwidth}
        \includegraphics[width=\textwidth]{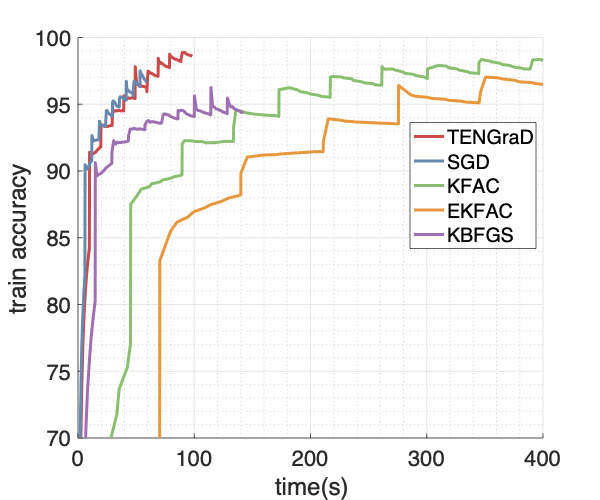}
         \caption{Fashion-MNIST}
        \label{fig:training_fashion}
    \end{subfigure}
    \begin{subfigure}[b]{0.32\textwidth}
        \includegraphics[width=\textwidth]{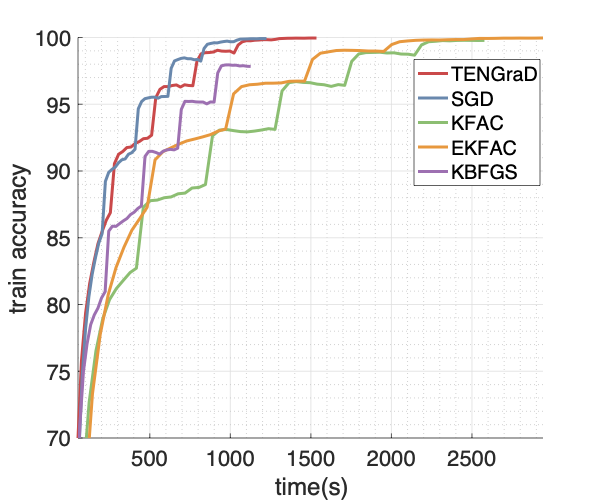}
         \caption{CIFAR-10}
        \label{fig:training_cifar10}
    \end{subfigure}
    \begin{subfigure}[b]{0.32\textwidth}
        \includegraphics[width=\textwidth]{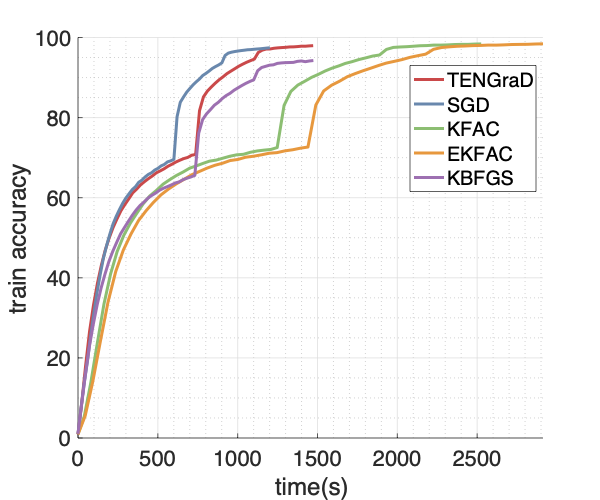}
         \caption{CIFAR-100}
        \label{fig:training_cifar100}
    \end{subfigure}
    \caption{ Comparing train accuracy between optimization methods for 
     a) Fashion-MNIST on 3C1F network. b) CIFAR-10 on Wide-ResNet network c) CIFAR-100 on Wide-ResNet Network.}\label{fig:training}
\end{figure*}

\begin{figure*}[h!]
    \centering
    \begin{subfigure}[b]{0.32\textwidth}
        \includegraphics[width=\textwidth]{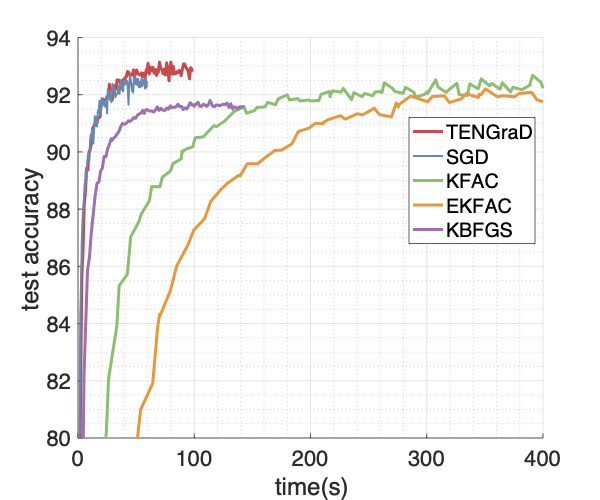}
         \caption{Fashion-MNIST}
        \label{fig:test_fashion}
    \end{subfigure}
    \begin{subfigure}[b]{0.32\textwidth}
        \includegraphics[width=\textwidth]{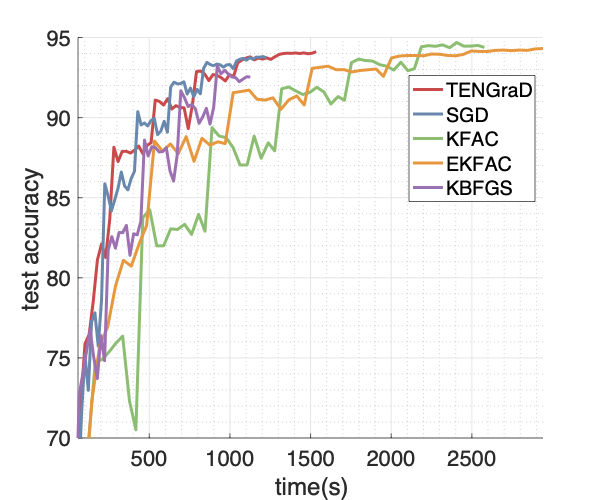}
         \caption{CIFAR-10}
        \label{fig:test_cifar10}
    \end{subfigure}
    \begin{subfigure}[b]{0.32\textwidth}
        \includegraphics[width=\textwidth]{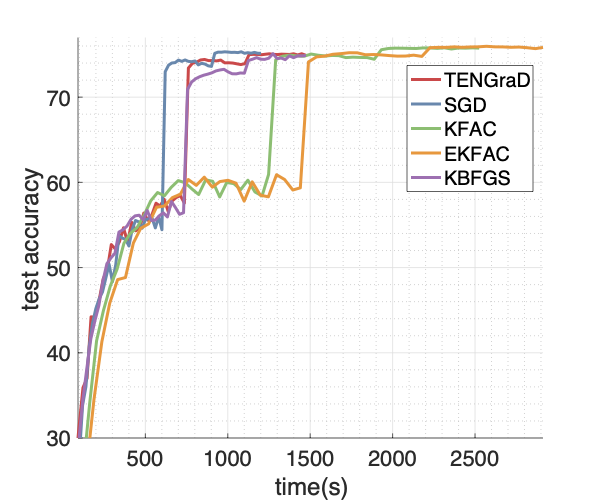}
         \caption{CIFAR-100}
        \label{fig:test_cifar100}
    \end{subfigure}
    \caption{ Comparing test accuracy between optimization methods for 
     a) Fashion-MNIST on 3C1F network. b) CIFAR-10 on Wide-ResNet network c) CIFAR-100 on Wide-ResNet Network.} \label{fig:test}
\end{figure*}

From the training loss plots in \autoref{fig:training}, TENGraD clearly outperforms all other approximate NGD methods in timing and train accuracy. Moreover, TENGraD performs better than SGD in both time and accuracy on Fashion-MNIST dataset, as shown in \autoref{fig:training_fashion},  and it has a competitive timing compared to SGD on both CIFAR-10 and CIFAR-100 as shown in \autoref{fig:training_cifar10} and \autoref{fig:training_cifar100}. The test accuracy reported in \autoref{fig:test} shows that TENGraD also generalizes well and achieves state-of-the-art accuracy on all benchmarks, and especially outperforms KBFGS with a margin of ~1.6\% . Moreover, TENGraD clearly outperforms SGD in both time and test accuracy on Fashion-MNIST, as shown in \autoref{fig:test_fashion}, with competitive results on CIFAR-10 and CIFAR-100.

To further demonstrate the performance of TENGraD, we conduct extensive experiments on state-of-the-art DNNs and report the time and test accuracy for the 7 configurations in \autoref{tab:results}. Our results show that TENGraD outperforms KFAC, EKFAC, and KBFGS in overall time in 7 out of 7 (7/7) experiments. Moreover, TENGraD achieves a better accuracy, with a margin between 0.4\%-1.41\%, in 6/7 experiments compared to KBFGS which is the fastest method among approximate Kronecker-factored NGD methods. Also, compared to SGD, TENGraD achieves a better timing for 3/7 experiments and a better test accuracy in 5/7.   In particular, TENGraD shows a 20\% improvement over SGD in time for the Fashion-MNIST dataset and a better timing for CIFAR-10 on Wide-ResNet and DenseNet.\looseness=-1
\vspace{-0.2cm}
\renewcommand{\tabcolsep}{3pt}
\begin{table}[h]
\centering
\caption{Top test accuracy and  time for state-of-the-art models and different natural gradient methods.}
 \label{tab:results}
\begin{tabular}{l|c|c|c|c|c|c}
\hline
\multirow{2}{*}{Models} &
 \multirow{2}{*}{Dataset} &
  \multicolumn{5}{c}{Time(s) [Accuracy\%]} \\ 
  \cline{3-7} 
% \hline
% \multicolumn{1}{c}{} 
&
   &
  SGD &
  TENGraD &
  KFAC &
  EKFAC &
  KBFGS \\ \hline
  3C1F &
  F-MNIST &
  52 [92.60] &
  \textbf{43} [93.15] &
  188 [92.2] &
  275 [92.01] &
  113 [92.12]\\ \hline
WideResNet &
  \begin{tabular}[c]{@{}c@{}}CIFAR-10\\ CIFAR-100\end{tabular} &
  \begin{tabular}[c]{@{}c@{}}1128 [94.01]\\ 1004 [75.79]\end{tabular} &
  \begin{tabular}[c]{@{}c@{}}\textbf{1108} [94.10]\\ \textbf{1124} [75.20]\end{tabular} &
  \begin{tabular}[c]{@{}c@{}}2454 [94.50]\\ 2419 [75.60]\end{tabular} &
  \begin{tabular}[c]{@{}c@{}}2936 [94.10]\\ 2766 [76.07]\end{tabular} &
  \begin{tabular}[c]{@{}c@{}}1132 [92.70]\\ 1151 [74.35]\end{tabular} \\ \hline
DenseNet &
  \begin{tabular}[c]{@{}c@{}}CIFAR-10\\ CIFAR-100\end{tabular} &
  \begin{tabular}[c]{@{}c@{}}2278 [93.52]\\ 1909 [72.10]\end{tabular} &
  \begin{tabular}[c]{@{}c@{}}\textbf{2077} [93.40]\\ \textbf{2002} [73.90]\end{tabular} &
  \begin{tabular}[c]{@{}c@{}}4292 [93.31]\\ 3542 [74.50]\end{tabular} &
  \begin{tabular}[c]{@{}c@{}}4820 [93.11]\\ 3788 [74.29]\end{tabular} &
  \begin{tabular}[c]{@{}c@{}}2412 [93.41]\\ 2292 [73.50]\end{tabular} \\ \hline
  MobileNetV2 &
  \begin{tabular}[c]{@{}c@{}}CIFAR-10\\ CIFAR-100\end{tabular} &
  \begin{tabular}[c]{@{}c@{}}1350 [92.70]\\ 1370 [72.78]\end{tabular} &
  \begin{tabular}[c]{@{}c@{}}\textbf{1841} [92.71]\\ \textbf{1406} [72.81]\end{tabular} &
  \begin{tabular}[c]{@{}c@{}}3221 [92.63]\\ 3152 [72.82]\end{tabular} &
  \begin{tabular}[c]{@{}c@{}}4304 [92.83]\\ 4687 [72.97]\end{tabular} &
  \begin{tabular}[c]{@{}c@{}}1728 [91.30]\\ 1695 [72.01]\end{tabular} \\ \hline
\end{tabular}
\end{table}

\begin{figure*}[h!]
    \centering
    \begin{subfigure}[b]{0.25\textwidth}
      \includegraphics[width=\textwidth]{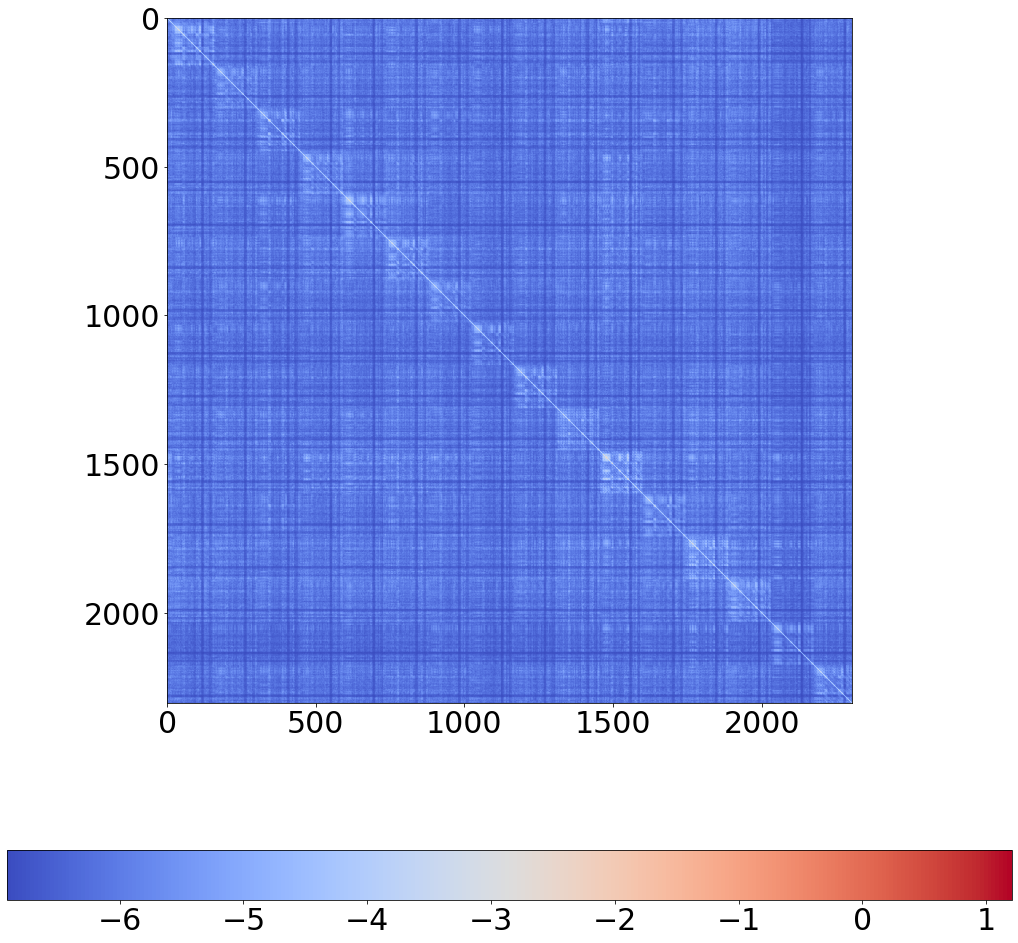}
         \caption{Exact NGD}
        \label{fig:exact_fisher_inv}
    \end{subfigure}
    \begin{subfigure}[b]{0.25\textwidth}
        \includegraphics[width=\textwidth]{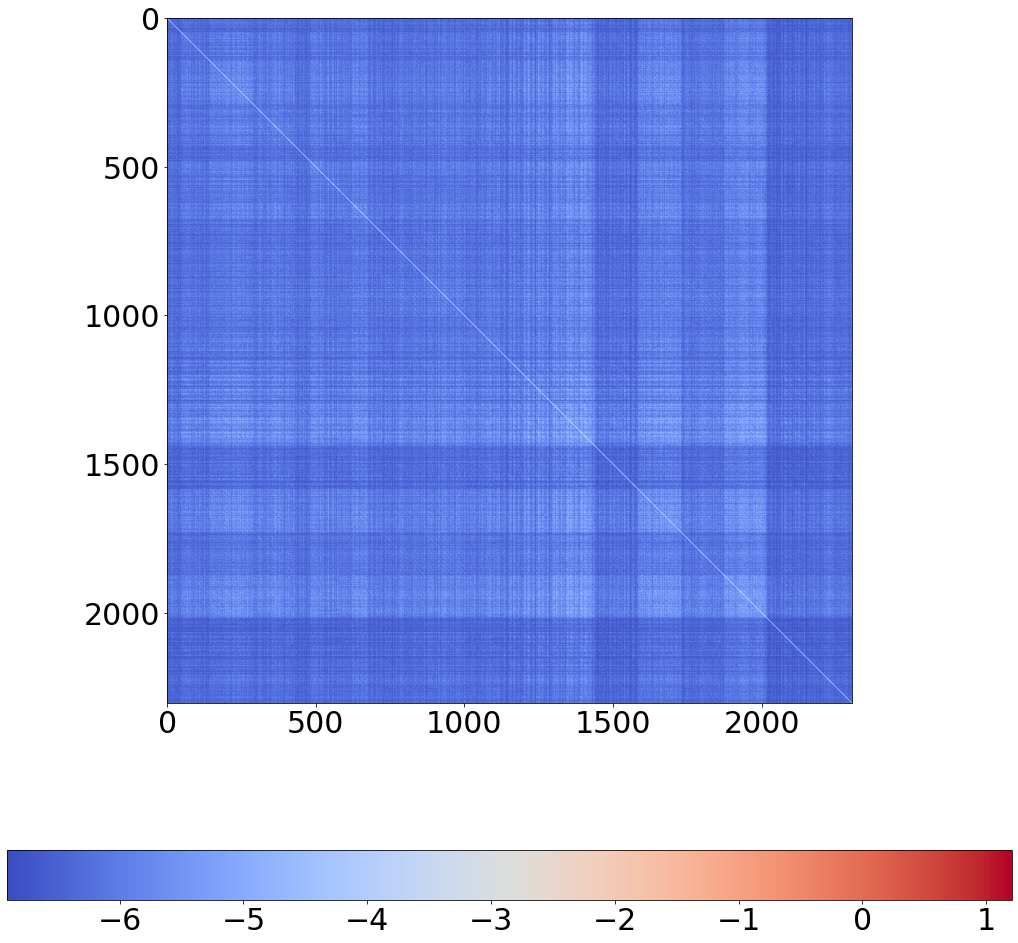}
         \caption{TENGraD}
        \label{fig:TENGraD_fisher_inv}
    \end{subfigure}
    \begin{subfigure}[b]{0.25\textwidth}
        \includegraphics[width=\textwidth]{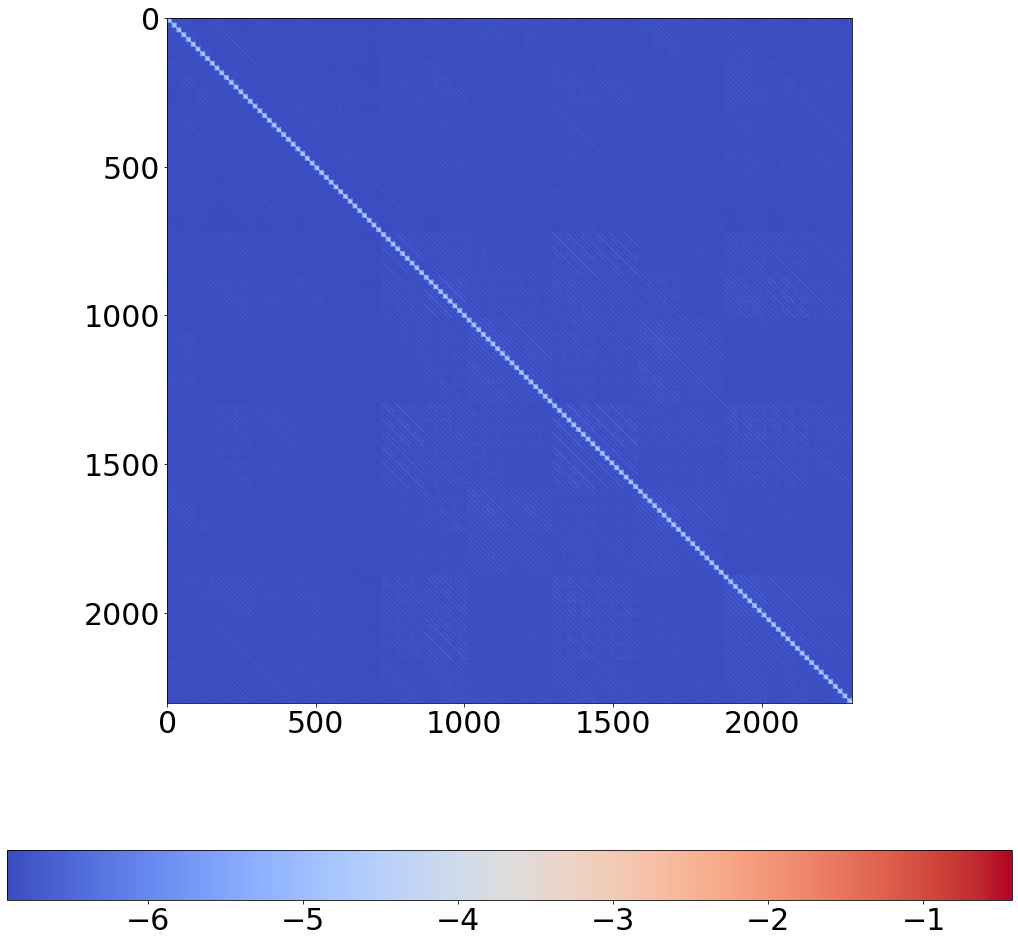}
         \caption{KFAC}
        \label{fig:kfac_fisher_inv}
    \end{subfigure}
    \caption{Structure of block inverses for the second convolution layer in 3C1F with Fashion-MNIST.  }\label{fig:approximation_quality}
\end{figure*}

To compare the quality of Fisher matrix approximation, we  compare the block-diagonal approximation  in TENGraD and KFAC with exact NGD. From \autoref{fig:approximation_quality}, TENGraD provides a similar approximation to the exact NGD by preserving the structure of the Fisher matrix. However, KFAC's approximation hardly captures the structure. Similar results for other layers are observed (see Appendix). %\bc{Lastly, we demonstrate the robustness by examining the train accuracy .....}

To demonstrate the robustness, we examine the train accuracy under various hyperparemeter (HP)
settings and show TENGraD is stable under a wide range for the HPs (see Appendix).
\vspace{-0.5cm}
\section{Conclusion}
\vspace{-0.2cm}
We propose a Time-Efficient NGD method, called TENGraD, that improves the overall time of second order methods with computationally  efficient factorization and reuse methods and prove linear convergence rates under certain assumptions.  Extension to convolutional and general layers is provided. Results indicate that TENGraD
outperforms or perform comparably to the state-of-the-art NGD methods as well as SGD. %A limitation of our approach is that it is asymptotic and requires number of samples to be large. 
Our work does not have a direct negative impact as it is mostly theoretical. \looseness=-1

\section*{Acknowledgements}
This work was supported in part
by NSERC Discovery Grants (RGPIN-06516, DGECR-00303), XSEDE, the
Canada Research Chairs program, and NSF grants CCF-1814888, NSF DMS-2053485, NSF DMS-1723085 and Office of Naval Research Award Number N00014-21-1-2244.
\bibliographystyle{unsrt}
\bibliography{main}

%%%%%%%%%%%%%%%%%%%%%%%%%%%%%%%%%%%%%%%%%%%%%%%%%%%%%%%%%%%%
\newpage
\appendix
% \tableofcontents
\newpage
\appendix
\section{Appendix}
\subsection{Outline}
The general form of the covariance factorization discussed in Section 3.2 is provided in Section \ref{sec:supp_general_from}. The convergence proof of Theorem 1 discussed in Section 4 is provided in Section \ref{sec:supp_convergence_tengrad}. The bounds for minimum eigenvalue $\lambda_0$ discussed in Section 4 is provided in \ref{sec:supp_bound_lam}. We provide the setup information in Section \ref{sec:supp_experiment_setup}. The plots that compare the quality of approximation of Fisher inverse blocks are provide in Section \ref{sec:supp_structure}. We provide the sensitivity to hyperparameters figures in Section \ref{sec:supp_sensitivity_hp}. For the error bars refer to Section \ref{sec:err_bars}.

% \tableofcontents
\subsection{General From of Covariance Factorization}
\label{sec:supp_general_from}
In this section, we extend the results obtained in  Section 3 and show that the covariance factorization and reuse method can be applied to an arbitrary layer. An arbitrary layer can operate on its input tensor in many ways depending on its underlying linear transformations. Typically these transformations operate differently on each tensor dimension. For example, a transformation such as convolution applies a specific filtering on spatial dimensions. Therefore, we need to distinguish between these dimensions depending on the operations performed on them. In the following, we categorize these dimensions into free, independent, and common dimensions.

\textbf{Tensor Dimensions}.
A \textit{common dimension} is the dimension of the unfolded input that is shared amongst input and output tensors, i.e. spatial dimensions. An \textit{independent dimension} is  the dimension of the unfolded input that is shared amongst input and the parameters, i.e. input channel $c_{in}$ and filter dimensions $k$. A \textit{free dimension} is
the dimension of the parameters that is shared amongst output and the parameters, i.e. output channel $c_{out}$.

% As an example, for a standard convolution layer, typically the dimensions represents batch $m$, input channel $C_{in}$, and spatial dimensions, i.e. width $W$ and height $H$.  Similarly, parameters of a convolution layer, i.e. filters, have dimensions input channel, output channel $C_out$ and spatial dimensions of filters , i.e. $W_f$ and $H_f$. 

% For convolutional layers, patches with the same size as filter, i.e. $W_f\times H_f$, are extracted from spatial dimensions, flatten into a row and stacked in a new Tensor.

% \textit{free mode}: a mode stays unchanged during the forward pass o

Now, we show that the Jacobian matrix $\bm{J}$ can be formed as a linear transformation amongst two tensors namely input and pre-activation derivatives which later is used to compute the Gram Jacobian $\bm{J}\bm{J}^\top$.

\begin{lem}
\label{lem:tensor_decomp}
Consider a layer with parameters   $\thetab_{v,i_1,...,i_k}$ that applies a linear transformation on input tensor $\mathcal{I}_{m,j_1,..,j_l}$ where $v$ and $m$ are the indices of free dimension and batch samples. Suppose that the output for a sample can be computed without requiring other samples, i.e. there is no dependency amongst samples in a batch. The pre-activation derivatives are given by tensor $\mathcal{G}_{m,v,t_1,...,t_p}$. We refer to set $\{t_1,...,t_p\}$ as common dimensions or $M_C$ and $\{i_1,...,i_k\}$ as independent dimensions or $M_I$. Then the Jacobian tensor can be written as:
\begin{equation}
    \label{eq:jacobian_tensor}
    \mathcal{J}_{m,v,M_I} = \sum_{M_C} \mathcal{G}_{m,v,M_C} \hat{\mathcal{I}}_{m,M_C,M_I}
\end{equation}
where tensor $\hat{I}$ is obtained by unfolding the input tensor $I$:
\begin{equation}
    \hat{\mathcal{I}}_{m,M_C,M_I} = \sum_{j_{\alpha} \in \Omega} \mathcal{I}_{m,j_1,...,j_l}
\end{equation}
where $\Omega$ depends on the underlying linear transformation.

\end{lem}

\textbf{Proof of \autoref{lem:tensor_decomp}}. Assume the arbitrary linear transformation as follow:

\begin{equation}
    \label{eq:linear_trans}
    \mathcal{O}_{m,v,t_1,...,t_p} = \sum_{j_\alpha\in \Omega , i_\beta \in \Gamma} \mathcal{I}_{m,j_1,...,j_p} \thetab_{v,i_1,...,i_k}
\end{equation}

Where $\Omega$ and $\Gamma$ define the underlying linear transformation such as the input and filter spatial dimensions in the convolutional layer. Now, we can rewrite \autoref{eq:linear_trans} by unfolding the input tensor. To do so, for each index $i_\beta \in \Gamma$ we can unfold the input tensor using the following:

\begin{equation}
    \hat{\mathcal{I}}_{m,t_1,...t_p,i_1,...,i_k} = \sum_{j_{\alpha} \in \Omega} \mathcal{I}_{m,j_1,...,j_l}
\end{equation}

The above equation unfolds the input along all the dimensions in the shared dimensions with the output tensor, i.e. $M_C$, and for all the dimensions in the parameter tensor. As a result, the output can be written as the following:

\begin{equation}
    \mathcal{O}_{m,v,M_C} = \sum_{M_I} \hat{\mathcal{I}}_{m,M_C,M_I} \thetab_{v,M_I}
\end{equation}

Using this result and the fact that the pre-activation derivatives tensor $\mathcal{G}$ has the same dimension as output tensor $\mathcal{O}$, we can obtain the Jacobian tensor $\mathcal{J}$ as stated in the \autoref{eq:jacobian_tensor}.

Lemma 1 shows that the Jacobian tensor $\mathcal{J}$ can be achieved with a linear transformation between the unfolded input tensor $\mathcal{\hat{I}}$ and the pre-activation derivative tensor $\mathcal{G}$.
Now, using this result we can extend the covariance factorization in TENGraD to an arbitrary layer.

\begin{thm}
\label{theorem:decomposition}
For a linear layer with input tensor $\mathcal{I}$ and pre-activation derivatives $\mathcal{G}$ defined in \autoref{lem:tensor_decomp}, the Gram matrix $ \bm{J} \bm{J}^\top$ is obtained by:
\begin{equation*}
    \bm{J} \bm{J}^\top = \sum_{M_C} \sum_{M'_C} C_1 \odot C_2
\end{equation*}
where $C_1$ and $C_2$ are the covariance tensors of unfolded input and pre-activation derivatives defined as:
\begin{equation*}
\begin{split}
    [C_1]_{m,m',M_C, M'_C} &=\sum_{M_I}\hat{\mathcal{I}}_{m,M_C,M_I}\hat{\mathcal{I}}_{m',M'_C,M_I} \\
    [C_2]_{m,m',M_c, M'_C} &= \sum_{v} \mathcal{G}_{m,v,M_C} \mathcal{G}_{m',v,M'_C}
\end{split}
\end{equation*}
\end{thm}

\textbf{Proof of \autoref{theorem:decomposition}}. Using \autoref{lem:tensor_decomp} we can write the Gram Jacobian matrix by expanding the common dimensions as follows:

\begin{equation*}
\begin{split}
       \bm{J}\bm{J}^\top &= \sum_v \sum_{M_I} \mathcal{J}_{m,v,M_I} \mathcal{J}_{m',v,M_I}\\
       &=\sum_{M_C} \sum_{M'_C}\sum_v \sum_{M_I} \mathcal{G}_{m,v,M_C}
       \mathcal{G}_{m',v,M'_C}
       \hat{\mathcal{I}}_{m,M_C,M_I}  \hat{\mathcal{I}}_{m',M'_C,M_I}\\
       &=\sum_{M_C} \sum_{M'_C} C_1 \odot C_2
\end{split}
\end{equation*}

As a result, the Gram Jacobian $\bm{J}\bm{J}^\top$ for any layer can be obtained by a reduction on the Hadamard product of two small covariance tensors. In the following Corollary we state the conditions such that \autoref{theorem:decomposition} improves the performance in practice by reducing the computation complexity.

\begin{corollary}
\label{cor:performance}
Suppose we show the size of common dimensions, independent dimensions and free dimensions by $N_C=\prod_{m_C \in M_C} m_c$, $N_I = \prod_{m_I \in M_I} m_I$ and $N_v$. Then \autoref{theorem:decomposition} reduces the complexity of Gram matrix computation from  
$O(m^2 N_I N_v + m N_I N_v N_C)$ to $O(m^2 {N_C}^2 (N_I + N_v))$.
\end{corollary}

\begin{remark}
\autoref{theorem:decomposition} applies to a wide range of layers in deep neural networks such as fully connected, convolution, and group convolution. It can also be applied to layers with nonlinear transformation as long as the non-linearity only applies to the input data such as layer normalization.  
\end{remark}

\subsection{Proofs of Section 4}

\subsubsection{Convergence Analysis of TENGraD}
\label{sec:supp_convergence_tengrad}
In this section, we provide the proof for Theorem 1. If Assumption 2 holds, we have $\norm{\bm{\hat{J}}(\thetab) - \bm{\hat{J}}(0)}_2 \leq \frac{C}{3}\lambda_{0} ^ {\frac{1}{2}}$. This result can help  bound the minimum eigenvalue of the Gram Jacobian at each iteration, stated in the following Lemma.

\begin{lem}
\label{lem:eigenvalue_G}
If $||\thetab(k) - \thetab(0)||_2 \leq 3\lambda_{0} ^ {\frac{-1}{2}} ||\bo{y}-\bo{u}(0)||_2 $, then we have $\lambda_{min}(\bm{\hat{G}}(k)) \geq \frac{4}{9} \lambda_0$.
\end{lem}
\textbf{Proof of \autoref{lem:eigenvalue_G}}. Based on the inequality $\sigma_{min}(A+B) \geq \sigma_{min}(A) - \sigma_{max}(B)$ and based on the fact that $\norm{\bm{\hat{J}}(\thetab) - \bm{\hat{J}}(0)}_2 \leq \frac{C}{3}\lambda_{0} ^ {\frac{1}{2}}$, we have:
\begin{equation*}
\begin{split}
 \sigma_{min}(\bm{\hat{J}}(k)) &\geq \sigma_{min}(\bm{\hat{J}}(0)) - || \bm{\hat{J}}(k) - \bm{\hat{J}}(0) ||_2\\
 &\geq \lambda_0^{\frac{1}{2}} - \frac{1}{3}\lambda_0^{\frac{1}{2}} = \frac{2}{3} \lambda_0^{\frac{1}{2}} \Rightarrow \lambda_{min}(\bm{\hat{G}}(k)) \geq \frac{4}{9} \lambda_0
\end{split}
\end{equation*}

We can now prove the linear convergence rate in Theorem 1. 

\textbf{Proof of Theorem 1}. 
We prove Theorem 1 by induction. Assume $|| \bo{u}(k) - \bo{y}||_2^2 \leq (1 - \eta)^k || \bo{u}(0) - \bo{y}||_2^2$. We can see that $\bm{J}^\top(k) = \bm{\hat{J}}^\top(k)  \bm{T}$ with $\bm{T}\in \mathbb{R}^{Ln\times n}$ and   $\bm{T} = [I_n,\dots,I_n]^\top$ where $I_n$ is the identity matrix of dimension $n$.  
\begin{align*}
\bm{u}(k+1) - \bm{u}(k) &= \bm{u}(\thetab(k) - \frac{\eta}{n} \left( \bm{\hat{F}}(k) + \alpha I \right) ^{-1}  \bm{J}(k)^\top (\bm{u}(k) - \bm{y}))    - \bm{u}(\thetab (k))\\
&= -\int_{s=0}^1  \inp[\Big]{\frac{\partial \bm{u}(\thetab(s))   }{\partial \thetab ^\top }}{\frac{\eta}{n}  \left( \bm{\hat{F}}(k) + \alpha I \right) ^{-1}  \bm{J}(k)^\top (\bm{u}(k) - \bm{y}))} ds \\
&= -\underbrace{\int_{s=0}^1  \inp[\Big]{\frac{\partial \bm{u}(\thetab(k))   }{\partial \thetab ^\top }}{\frac{\eta}{n}  \left( \bm{\hat{F}}(k) + \alpha I \right) ^{-1}  \bm{J}(k)^\top (\bm{u}(k) - \bm{y}))} ds}_{\circled{1}}\\
&+ \underbrace{\int_{s=0}^1  \inp[\Big]{\frac{\partial \bm{u}(\thetab(k))   }{\partial \thetab ^\top } - \frac{\partial \bm{u}(\thetab(s))   }{\partial \thetab ^\top }}{\frac{\eta}{n}  \left( \bm{\hat{F}}(k) + \alpha I \right) ^{-1}  \bm{J}(k)^\top (\bm{u}(k) - \bm{y}))} ds}_{\circled{2}}
\end{align*}
where $\thetab(s) = \thetab(k) - \frac{s\eta}{n} \left( \bm{\hat{F}}(k) + \alpha I \right) ^{-1}  \bm{J}(k)^\top (\bm{u}(k) - \bm{y})) $. Now, we bound each term separately:
\begin{align}% \begin{split}
\circled{1} &= -\frac{\eta}{n}  \bm{J}(k) \left( \bm{\hat{F}}(k) + \alpha I \right) ^{-1}  \bm{J}(k)^\top (\bm{u}(k) - \bm{y}))\label{eq:proof_1}\\
% \end{split}
% \end{equation}
% \begin{align}
% \begin{split}
||\circled{2}||_2 &\leq \frac{\eta}{n} \norm{ \int_{s=0}^1 \bm{J}(\thetab(s)) - \bm{J}(\thetab(k)) ds}_2 \norm{\left( \bm{\hat{F}}(k) + \alpha I \right) ^{-1}  \bm{J}(k)^\top (\bm{u}(k) - \bm{y}))}_2\nonumber\\
&\overset{(i)}{\leq} \frac{\eta 2C}{3n} \lambda_0^{\frac{1}{2}} \norm{\left( \frac{1}{n} \bm{\hat{J}}^\top \bm{\hat{J}} + \alpha I \right) ^{-1}  \bm{\hat{J}}^\top \bm{T} (\bm{u}(k) - \bm{y}))}_2 \nonumber\\
&\overset{(ii)}{\leq} \frac{\eta 2C}{3} \lambda_0^{\frac{1}{2}} \frac{\lambda_{min}(\bm{\hat{G}}(k))^{\frac{1}{2}}}{\lambda_{min}(\bm{\hat{G}}(k)) + n \alpha} \norm{\bm{T} (\bm{u}(k) - \bm{y}))}_2 \nonumber\\
&\overset{(iii)}{\leq}\frac{\eta C \lambda_0 \sqrt{L}}{\lambda_0 + \frac{9}{4}n \alpha}  \norm{(\bm{u}(k) - \bm{y}))}_2 \label{eq:proof_2}
% \end{split}
\end{align}
We use Assumption 2 for the second inequality (i) which implies:
\begin{align*}
% \begin{split}
    \norm{ \int_{s=0}^1 \bm{J}(\thetab(s)) - \bm{J}(\thetab(k)) ds}_2 &\leq \norm{ \bm{J}(\thetab(k)) - \bm{J}(\thetab(0)) }_2 + \norm{ \bm{J}(\thetab(k+1)) - \bm{J}(\thetab(0)) }_2\nonumber\\
    &\leq \frac{2C}{3} \lambda_0^{\frac{1}{2}}.
% \end{split}
\end{align*}
The inequality (ii) follows from
\autoref{lem:eigenvalue_G} and in the last inequality (iii) we have used the fact that $\norm{\bm{T} (\bm{u}(k) - \bm{y}))}_2 = \sqrt{L} \norm{ (\bm{u}(k) - \bm{y}))}_2$.
Finally, we are ready to prove the convergence:
\begin{align*}
% \begin{split}
|| \bm{u}(k+1) - \bm{y}||_2^2
&\leq || \bm{u}(k) - \bm{y}||_2^2 -2 \left( \bm{y} - \bm{u}(k) \right) ^\top  \left( \bm{u}(k+1)- \bm{u}(k) \right) + ||\bm{u}(k+1)- \bm{u}(k) ||_2^2\\
&\leq || \bm{u}(k) - \bm{y}||_2^2 -\frac{2\eta}{n} \underbrace{\left( \bm{y} - \bm{u}(k) \right) ^\top  \bm{J}(k) \left( \bm{\hat{F}}(k) + \alpha I \right) ^{-1}  \bm{J}(k)^\top \left( \bm{y} - \bm{u}(k) \right) }_{\circled{A}}\\
&\quad\quad\quad+ \frac{2\eta C \lambda_0 \sqrt{L}}{\lambda_0 + \frac{9}{4}n \alpha}  \norm{(\bm{u}(k) - \bm{y}))}_2^2 + \underbrace{||\bm{u}(k+1)- \bm{u}(k) ||_2^2}_{\circled{B}}\\
&\leq || \bm{u}(k) - \bm{y}||_2^2 - \frac{2\eta L \lambda_0}{\lambda_0 + \frac{9}{4}n\alpha}  || \bm{u}(k) - \bm{y}||_2^2 + \frac{2\eta C \lambda_0 \sqrt{L}}{\lambda_0 + \frac{9}{4}n \alpha}  \norm{(\bm{u}(k) - \bm{y}))}_2^2 \\
&\quad\quad\quad+ \eta^2 \left(L + \frac{ C \lambda_0 \sqrt{L}}{\lambda_0 + \frac{9}{4}n \alpha}\right)^2\norm{(\bm{u}(k) - \bm{y}))}_2^2 \\
&\leq (1-\eta) \norm{(\bm{u}(k) - \bm{y}))}_2^2,
% \end{split}
\end{align*}
where in the last inequality we use the assumption that $\eta < \frac{2L\gamma -1 - 2C\sqrt{L}\gamma }{(L +C\sqrt{L}\gamma)^2 }$ where $\gamma = \frac{\lambda_0}{\lambda_0 + \frac{9}{4}n\alpha}$. Part $\circled{A}$ follows from \autoref{lem:eigenvalue_G}:
\begin{equation*}
    \begin{split}
        \circled{A} &\leq nL \Vert \bm{u}(k) - \bm{y} \Vert_2^2 \frac{\lambda_{min}\bm{\hat{G}}(k)}{\lambda_{min}\bm{\hat{G}}(k) + n\alpha} \leq \frac{nL\lambda_0}{\lambda_0 + \frac{9}{4}n\alpha} \Vert \bm{u}(k) - \bm{y} \Vert_2^2,
    \end{split}
\end{equation*}
Part $\circled{B}$, on the other hand, is implied by Equation \eqref{eq:proof_1}
and Inequality \eqref{eq:proof_2}.

\subsubsection{Bounding $\lambda_0$}
\label{sec:supp_bound_lam}
The results \cite{nguyen2020tight} are not directly applicable to our setting as we work with block-diagonal matrices. In the following, we introduce the assumptions previously made in 
\cite{nguyen2020tight} and adapt their proofs to our setting with some modifications for providing a bound on the smallest eigenvalue of the Gram matrix. %\cite{ngr bound on the smallest eigenvalue 
%mihete that in practice scaling property can be satisfied by pre-processing the data and the log-Sobolev inequality holds for the most of the distributions that are assumed on $P_X$ in practice, such as standard Gaussian distribution
%Therefore we make the following assumptions on $P_X$ previously made in \cite{nguyen2020tight} and adapt their results to our settings with minor modifications, the end result is given in \ref{lem: SmallEigenValue} below.
\begin{assumption}[Scaling]\label{Assump: DataScaling} The distribution of the data $P_X$ satisfies the conditions: $\mathbb{E}_{P_X}[\Vert x \Vert_2]=\Theta(\sqrt{d})$, $\mathbb{E}_{P_X}[\Vert x \Vert^2_2]=\Theta(d)$, and $\mathbb{E}_{P_X}[\Vert x-\mathbb{E}_{P_X}[x]\Vert^2_2]=\Omega (d)$.
\end{assumption}
\begin{assumption}[Lipschitz Concentration]\label{Assump: Lib_Concentration} For every Lipschitz continuous function $f$ with Libschitz constant $\Vert f\Vert_{Lip}$, the random variable $f(x)$ is Sub-Gaussian. That is there exists an absolute constant $c>0$ such that for all $t>0$, we have $\mathbb{P}\{|f(x)-\mathbb{E}_{P_X}[f(x)]|>t\}\leq 2\exp\left\{-\frac{ct^2}{\Vert f \Vert_{Lip}^2}\right\}
$.
\end{assumption}
 Assumption \ref{Assump: Lib_Concentration} is satisfied for the family of distributions satisfying log Sobolev inequality with a dimension independent constant \cite{nguyen2020tight}. Under these assumptions on $P_X$, the authors of \cite{nguyen2020tight} have provided tight bounds on the smallest and largest eigenvalues of empirical Neural Tangent Kernel $\bar{K}^{(L)}:=\sum_{l=1}^{L}\bm{J}_{l}(k)\bm{J}_l(k)^\top$ by studying the spectrum of $J_{l}(k)J_l(k)^\top$ for each layer $l$. Therefore, their result translates to the spectrum of $\hat{\bm{J}}(k)\hat{\bm{J}}(k)^\top$ which is given by eigenvalues of $J_l(k) J_l(k)^\top $. We present the lower bound on $\lambda_0$ suggested by \cite[Theorem 4.1]{nguyen2020tight} at Lemma \ref{lem: SmallEigenValue} below and provide its derivation for the sake of completeness. 
\begin{lem}\label{lem: SmallEigenValue}
Consider a L-layer ReLU network with single output. Let weights of the network at each layer $l\in[L]$ be $W_{l}\in\mathbb{R}^{d_{l-1}\times d_l}$ and initialized as $(W_l)_{ij} \sim \mathcal{N}(0,\beta_l^2)$ for $l\in[L]$. Assume that data points $\{x_i,y_i\}$ are i.i.d. and drawn from a distribution $P_X$ which satisfies the Assumptions \ref{Assump: DataScaling} and \ref{Assump: Lib_Concentration}. Fix any $\delta>0$, and any even integer $r\geq 2$. Let $\mu_r(\sigma)$ be r-th Hermite coefficient of ReLU activation function $\sigma$ given as 
$$
\mu_r(\sigma)=\frac{1}{\sqrt{2\pi}}(-1)^{\frac{r-2}{2}}\frac{(r-3)}{\sqrt{r!}}.
$$ 
Suppose the number of neurons at each layer $l\in [L-1]$ satisfies the conditions
\begin{align*}
d_l=\Omega\left(N\log(N)\log(\frac{N}{\delta})\right) \;\; \&\;\;  \prod_{j=1}^{l-2}\log(d_j)=o(\min_{j\in[0,l-1]}d_j),
\end{align*}
then the following inequality holds  for all $k\geq 0$
$$
\lambda_{min}(\hat{\bm{J}}(k)\hat{\bm{J}}(k)^\top) \geq \min_{l\in [L]} \left\{\mu_r(\sigma)^2 \Omega\left(d\prod_{j=1}^{L-1}d_j\prod_{j=1,j\neq l}^{L}\beta_j^{2}\right),\lambda_{\min}(XX^\top)\Omega\left( \prod_{l=1}^{L-1}d_l \prod_{l=2}^{L}\beta_l^2\right) \right\},
$$
with probability at least 
$$
1-\delta - N^2 \exp\left(-\Omega\left( \frac{\min_{l'\in[0,l-1]}n_l}{N^{2/(r-0.1)\prod_{l'=1}^{l-2}\log(d_{l'})}}\right)\right)-N\sum_{l'=1}^{l-1}\exp(-\Omega(d_{l'}))-N\exp(-\Omega(d)).
$$ 
% $$
% 1-\delta-\sum_{k=1}^{L-1}N^2 \exp\left\{ -\Omega\left( \frac{\min_{l\in [0,k-1]}n_l}{N^{2/(r-0.1)\prod_{l=1}^{k-2}}\log(n_l)} \right)\right\}-N\sum_{l=1}^{L-1}\exp\left\{-\Omega(n_l)\right\}-N\exp(-\Omega(d)).
% $$
\end{lem} 
\textbf{Proof of Lemma \ref{lem: SmallEigenValue}}
In our analysis, we consider a L-layer ReLU network whose weights and biases are given by matrices $W_{l}\in\mathbb{R}^{d_{l-1}\times d_l}$ and each feature map $f_l:\mathbb{R}^{d}\rightarrow \mathbb{R}^{d_l}$ is defined as  
 \begin{equation*}
 f_l(x)=
 \begin{cases} 
 x, & \text{ if } l=0\\
 \sigma(W_l^\top f_{l-1}), & \text{ if } l\in [L-1]\\
 W_L^\top f_{L-1}. & \text{ if } l=L
 \end{cases}
 \end{equation*} 
 
where the Fisher-block matrix is written as $\bm{\hat{F}}(\bm{W}(k))=\bm{\hat{J}}(k) \bm{\hat{J}}(k)^\top$ with 
\begin{equation*}
% \label{eq:block_jacobian}
   \bm{\hat{J}}(k) =  \begin{bmatrix}
  \bm{J}_{1}(k) & 0 & \dots & 0\\ 
  0 & \bm{J}_{2}(k) & \dots & 0 \\
  \vdots & \ddots &  & \vdots  \\
  0 & \dots & 0 & \bm{J}_{L}(k) \\
\end{bmatrix} 
\end{equation*}

and each $\bm{J}_{l}(k)$ is the Jacobian of the network w.r.t the parameters in the layer $l$, i.e. $\bm{J}_{l}(k)=\frac{\partial \bm{f}(\bm{W}(k))}{\partial \text{vec}(W_l)}$. The result follows from \cite[Theorem 4.1]{nguyen2020tight}, but for completeness we provide it here. 
Let $\mathcal{O}_{l,j}(x)$ be the pre-activation neuron, $\sigma(.)$ be the ReLU function, and $\Sigma_{l}(x)=\text{Diag}\left(\sigma'(\mathcal{O}_{l,j}(x))\vert_{j=1}^{d_l} \right)\in \mathbb{R}^{d_l\times d_l}$. We can write the following equation for each of the layer $l$:
$$
\left(\frac{\partial F_L}{\partial \text{vec}(W_l)} \right)\left(\frac{\partial F_L}{\partial \text{vec}(W_l)}\right)^\top=\bm{f}_{l-1}\bm{f}_{l-1}^\top \circ B_{l}B^\top_{l} 
$$
where $\bm{f}_{l}=[\frac{\partial f_l(\bm{W};x_1)}{\partial \text{vec}(W_l)},....,\frac{\partial f_l(\bm{W};x_n)}{\partial \text{vec}(W_l)}]^\top$ and $B_{l}\in\mathbb{R}^{N\times N}$ is a matrix whose i-th row is given as
\begin{align*}
    [B_{l}]_{i:}=
    \begin{cases} 
    \Sigma_{l}(x_i)\left( \prod^{l-1}_{l'=l+1}W_{l'}\Sigma_{l'}(x_i)\right)W_L, & \; l \in [0,L-2]\\
    \Sigma_{L-1}(x_i)W_L, & \; l=L-1\\
    \frac{1}{\sqrt{N}}1_N, & \; l=L
    \end{cases} 
\end{align*}
The result relies on the inequality 
$$
\lambda_{\min}(F_{l-1}F_{l-1}^\top \circ B_{l}B_l^\top) \geq \lambda_{\min}(F_{l-1}F_{l-1}^\top) \min_{i\in \{1,..,N\}}\Vert [B_{l}]_{i:}\Vert_2^2.
$$
The Lemma 4.3 given at \cite{nguyen2020tight} yields that for any layer $l\in [1,...,L-2]$ and $x\sim P_X$ the following holds
$$
\left\Vert \Sigma_{l+1}(x)\left(\prod_{l'=l+2}^{L-1}W_{l'}\Sigma_{l'}(x)W_L\right)\right\Vert_2^2=\Theta\left(\beta_L^2d_{l+1}\prod_{l'=l+2}^{L-1}d_{l'}\beta_{l'}^2\right)
$$
w.p. at least $1-\sum_{l=1}^{L-1}\exp\{-\Omega(d_l)\}-\exp(-\Omega(d))$ assuming that the product term $\prod_{l'='+2}^{L-1}(.)$ is inactive for $l=L-2$. The Theorem 5.1 at \cite{nguyen2020tight} provides the lower bound on $\lambda_{\min}(F_lF_l^\top)$ which depends on $\mu_r(\sigma)$ in addition to problem parameters $d,d_1,...,d_l$. 
If we fix $l\in \{1,..,L-1\}$ and any integer $r>0$, then for $\delta>0$ and the $d_l$ satisfying  $d_l=\Omega\left(N \log(N)\log(\frac{N}{\delta}) \right)$ and $\prod_{l'=1}^{l-2}\log(d_{l'})=o(\min_{l'\in[0, l-1]}d_{l'})$, the smallest singular value of $F_l$ satisfies the inequality 
$$
\sigma_{\min}(F_l)^2 \geq \mu_r(\sigma)^{2}\Omega\left( d\prod_{l'=1}^{l}d_{l'}\beta_{l'}\right)
$$
w.p. at least 
$$
1-\delta - N^2 \exp\left(-\Omega\left( \frac{\min_{l'\in[0,l-1]}n_{l'}}{N^{2/(r-0.1)\prod_{l'=1}^{l-2}\log(d_{l'})}}\right)\right)-N\sum_{l'=1}^{l-1}\exp(-\Omega(d_{l'}))-N\exp(-\Omega(d)).
$$
% The r-th Hermite polynomial $\mu_r(\sigma)$ of ReLU function $\sigma$ can be computed for any even integer $r\geq 2$ as 
% $$
% \mu_r(\sigma)=\frac{1}{\sqrt{2\pi}}(-1)^{\frac{r-2}{2}}\frac{(r-3)}{\sqrt{r!}}.
% $$ 
% Combining these findings yields the desired result. 

\subsection{Experiments}
This section provides additional details on experiments and extra results. First, we provide details of the DNN architectures used in the experiments and then the training process and finally, we show additional results.

\subsubsection{Experimental Setup}
\label{sec:supp_experiment_setup}
\subsubsubsection{DNN Architectures}
For the \verb+Fashion-MNIST+ benchmark, we use a model with three convolutional layers and one fully connected layer. The first convolutional layer has one input channel and 128 output channels. The second and third layers have 128 in both input and output channels. The padding and stride are set to 1 and the kernel size is 3. The fully connected layer has an input of  $9^2 \times 128$ and  output of size  $500$. All the layers are followed by ReLU activation functions. Finally, there is a fully connected layer with $10$ outputs and no activation. For \verb+CIFAR-10+ and \verb+CIFAR-100+ we use the \verb+DenseNet+ \cite{huang2017densely}, \verb+WideResNet+ \cite{zagoruyko2016wide} and \verb+MobileNetV2+ \cite{sandler2018mobilenetv2} architectures. For \verb+WideResNet+ model, we use depth=28 and a widen factor of 4. For \verb+DenseNet+ we set depth to 19 and choose a growth rate of 100. For \verb+MobileNetV2+ we use the default model. 

\subsubsubsection{Training Process}
In all of the experiments, we use a batch size of 128 samples which is widely used in practice. We also apply momentum to all optimization methods since, from observation, it improves the training time and accuracy for all methods. To avoid over-fitting, we apply a standard weight decay where the weight decay parameter is tuned. For \verb+Fashion-MNIST+ we set the number of epochs to 10 and for \verb+CIFAR-10+ and \verb+CIFAR-100+, we use 60 epochs.
For NGD methods we use a frequency of inversion of 100 to amortize the cost of updates. For EKFAC we set the scaling frequency to 20.

\textit{Learning rate strategy}. For \verb+Fashion-MNIST+  we fix the learning rate during training. For \verb+CIFAR-10+, a decaying learning rate strategy is used to achieve state-of-the-art accuracy which is \bc{\sout{a}} typical in practice. We decay the learning rate by half after every 10 epochs. For \verb+CIFAR-100+, we decay the learning rate by $0.1$ after epochs 30 and 45. 

\textit{Tuning parameters}. We tune the parameter learning rate, weight decay, and  dampening and  use a grid search. The range of learning rate is $\{0.001, 0.003, 0.01, 0.03, 0.1, 0.3\}$. The range for the weight decay is $\{0.001, 0.003, 0.01, 0.3, 0.1, 0.3\}$. The range of damping parameter is $\{0.001, 0.003, 0.01, 0.03, 0.1, 0.2, 0.3\}$. The momentum is set to 0.9. \autoref{tab:cifar10_tuned}, \autoref{tab:cifar100_tuned} and \autoref{tab:fashionmnist_tuned} provide the tuned values for the parameters of each experiment.

\begin{table}[h!]
\caption{Tuned parameters (learning rate, weight decay and damping) for the Fashion-MNIST experiments.}
\label{tab:fashionmnist_tuned}
\begin{tabular}{l|l|l|l|l|l|}
\cline{2-6}
                           & \multicolumn{1}{c|}{TENGraD} & \multicolumn{1}{c|}{KFAC} & \multicolumn{1}{c|}{EKFAC} & \multicolumn{1}{c|}{KBFGS} & \multicolumn{1}{c|}{SGD} \\ \hline
\multicolumn{1}{|l|}{3C1F} & 0.003, 0.001, 0.1            & 0.003, 0.001, 0.1         & 0.001, 0.003, 0.03         & 0.03, 0.01, 0.01           & 0.03, 0.001              \\ \hline
\end{tabular}
\end{table}

\begin{table}[h!]
\caption{Tuned parameters (learning rate, weight decay and damping) for the CIFAR-10 experiments.}
\label{tab:cifar10_tuned}
\begin{tabular}{l|l|l|l|l|l|}
\cline{2-6}
 & \multicolumn{1}{c|}{TENGraD} & \multicolumn{1}{c|}{KFAC} & \multicolumn{1}{c|}{EKFAC} & \multicolumn{1}{c|}{KBFGS} & \multicolumn{1}{c|}{SGD} \\ \hline
\multicolumn{1}{|l|}{WideResNet}  & 0.03, 0.003, 0.1 & 0.03, 0.003, 0.03 & 0.01, 0.003, 0.03 & 0.03, 0.003, 0.03 & 0.01, 0.003 \\ \hline
\multicolumn{1}{|l|}{DenseNet}    & 0.1, 0.003, 0.2 & 0.01, 0.003, 0.03 & 0.01, 0.003, 0.01 & 0.01, 0.003, 0.01 & 0.03, 0.003 \\ \hline
\multicolumn{1}{|l|}{MobileNetV2} & 0.01, 0.01, 0.2  & 0.01, 0.003, 0.1  & 0.01, 0.003, 0.1  & 0.01, 0.03, 0.03  & 0.01, 0.003 \\ \hline
\end{tabular}
\end{table}

\begin{table}[h!]
\caption{Tuned parameters (learning rate, weight decay and damping) for the CIFAR-100 experiments.}
\label{tab:cifar100_tuned}
\begin{tabular}{l|l|l|l|l|l|}
\cline{2-6}
 & \multicolumn{1}{c|}{TENGraD} & \multicolumn{1}{c|}{KFAC} & \multicolumn{1}{c|}{EKFAC} & \multicolumn{1}{c|}{KBFGS} & \multicolumn{1}{c|}{SGD} \\ \hline
\multicolumn{1}{|l|}{WideResNet}  & 0.01, 0.03, 0.3 & 0.003, 0.01, 0.03 & 0.003, 0.01, 0.03 & 0.003, 0.01, 0.01 & 0.01, 0.003 \\ \hline
\multicolumn{1}{|l|}{DenseNet}    & 0.1, 0.003, 0.1 & 0.01, 0.003, 0.03 & 0.01, 0.003, 0.03 & 0.01, 0.003, 0.03 & 0.01, 0.003 \\ \hline
\multicolumn{1}{|l|}{MobileNetV2} & 0.01, 0.01, 0.3 & 0.01, 0.003, 0.03 & 0.01, 0.003, 0.03 & 0.03, 0.01, 0.03  & 0.01, 0.003 \\ \hline
\end{tabular}
\end{table}

\subsubsection{Structure of Block Inverses}
\label{sec:supp_structure}
\begin{figure*}[h!]
    \centering
    \begin{subfigure}[b]{0.25\textwidth}
      \includegraphics[width=\textwidth]{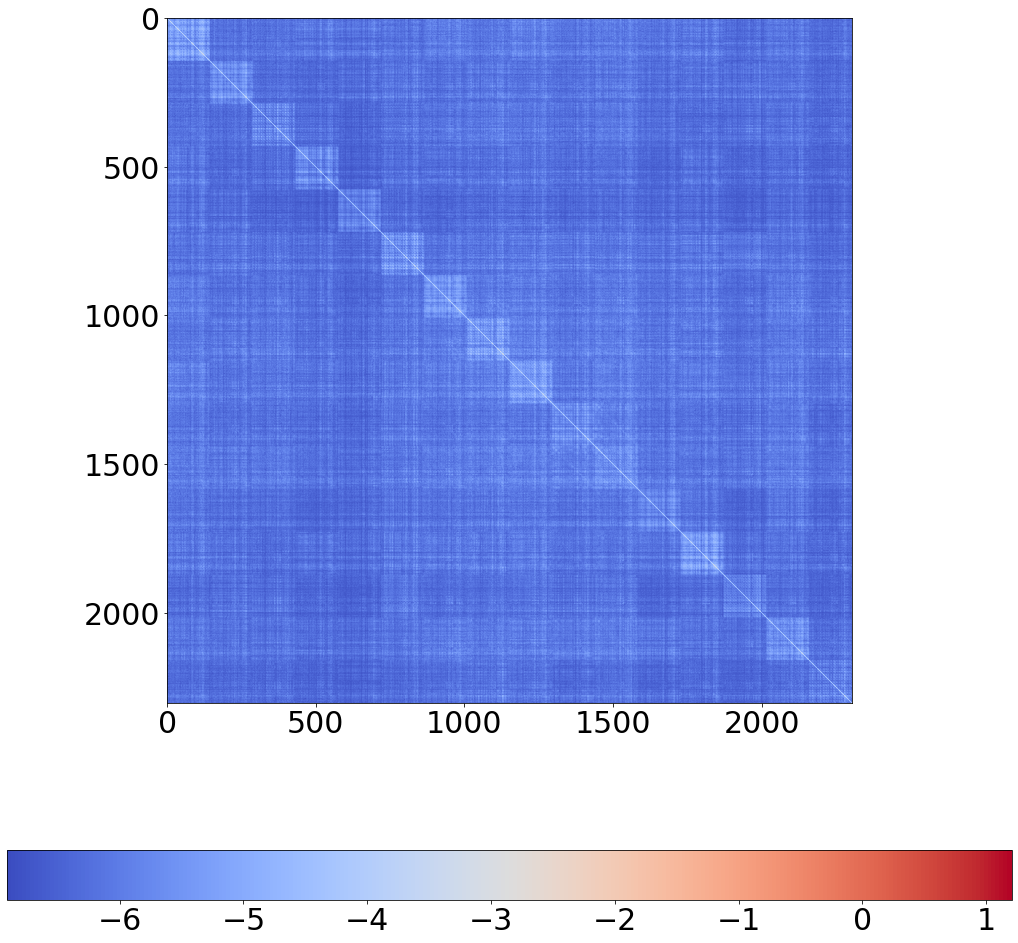}
         \caption{Exact NGD}
        \label{fig:exact_fisher_inv_3c}
    \end{subfigure}
    \begin{subfigure}[b]{0.25\textwidth}
        \includegraphics[width=\textwidth]{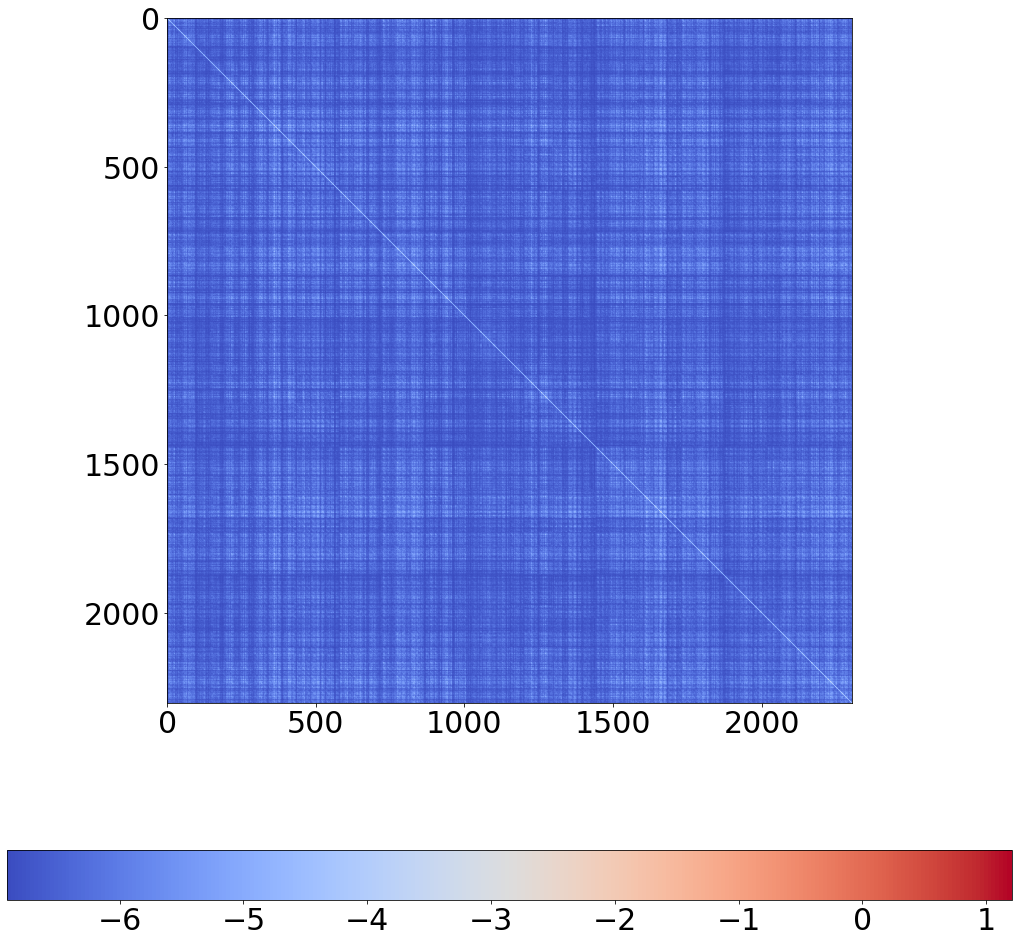}
         \caption{TENGraD}
        \label{fig:TENGraD_fisher_inv_3c}
    \end{subfigure}
    \begin{subfigure}[b]{0.25\textwidth}
        \includegraphics[width=\textwidth]{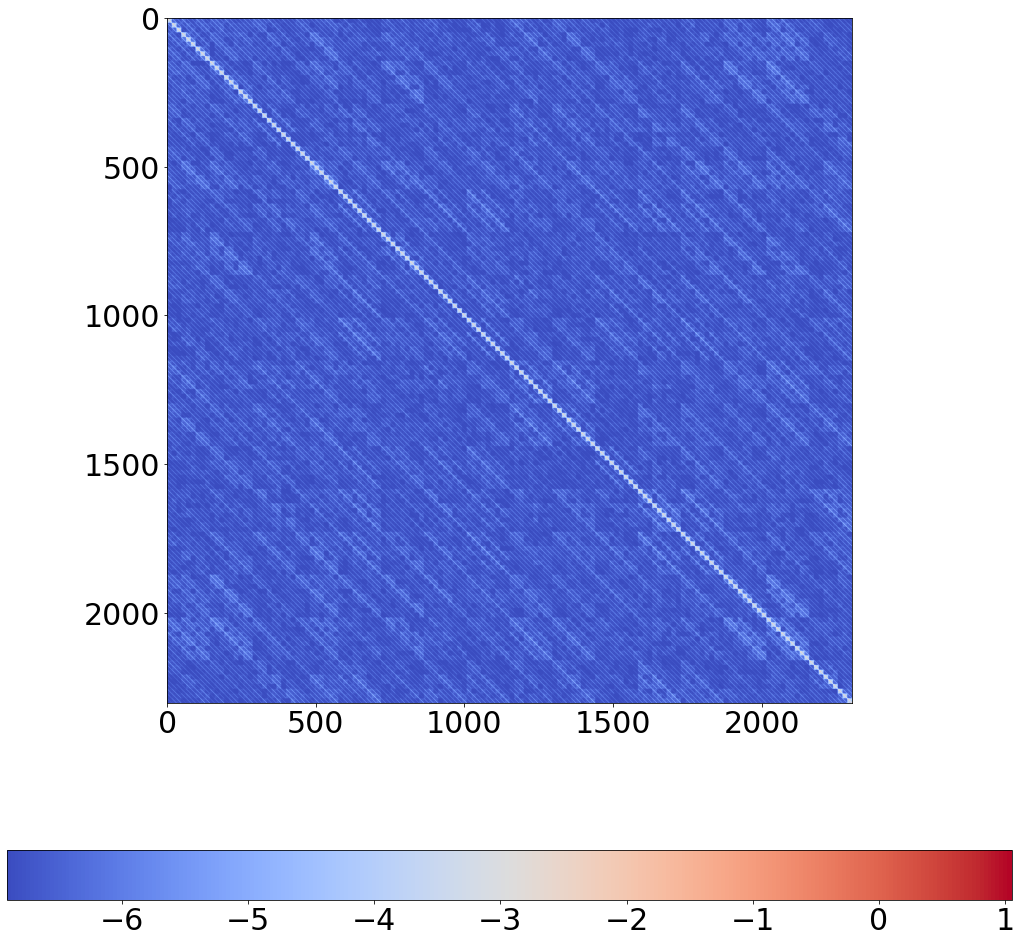}
         \caption{KFAC}
        \label{fig:kfac_fisher_inv_3c}
    \end{subfigure}
    \caption{Structure of block inverses for the third convolution layer in 3C1F with Fashion-MNIST.  }\label{fig:approximation_quality_3c}
\end{figure*}

\begin{figure*}[h!]
    \centering
    \begin{subfigure}[b]{0.25\textwidth}
      \includegraphics[width=\textwidth]{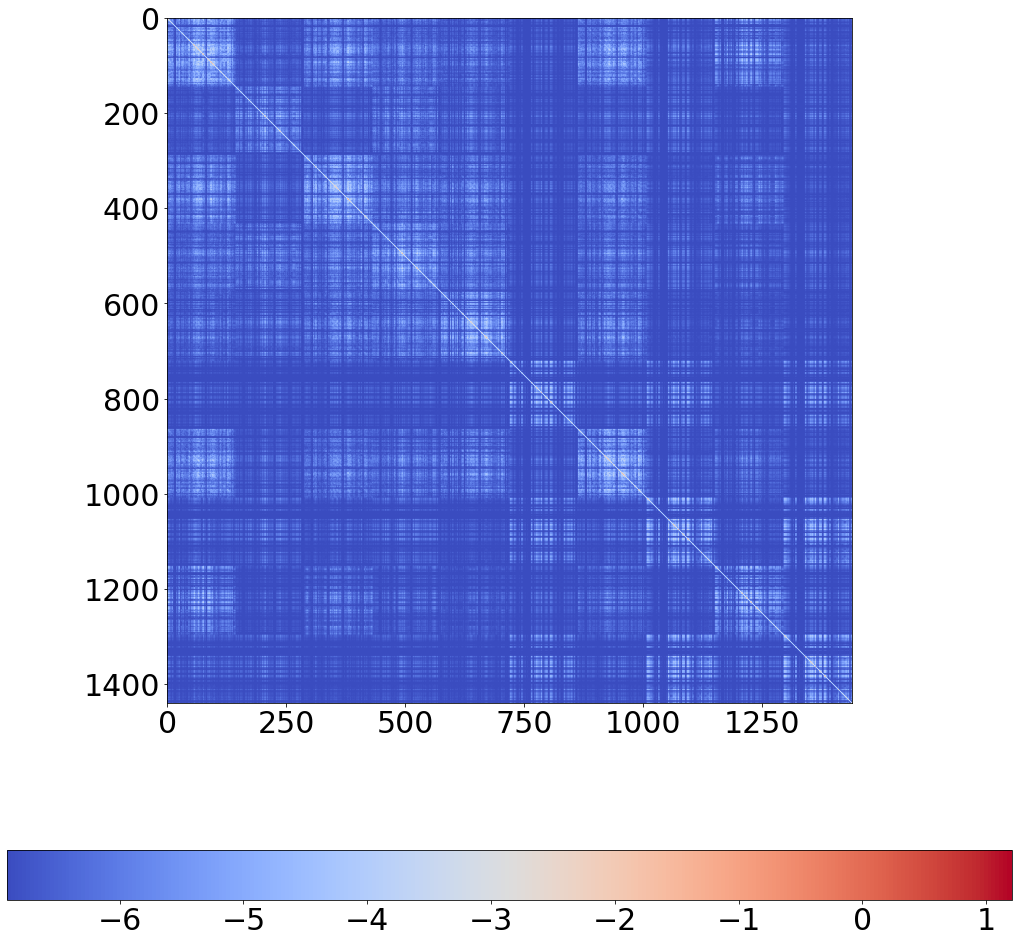}
         \caption{Exact NGD}
        \label{fig:exact_fisher_inv_l}
    \end{subfigure}
    \begin{subfigure}[b]{0.25\textwidth}
        \includegraphics[width=\textwidth]{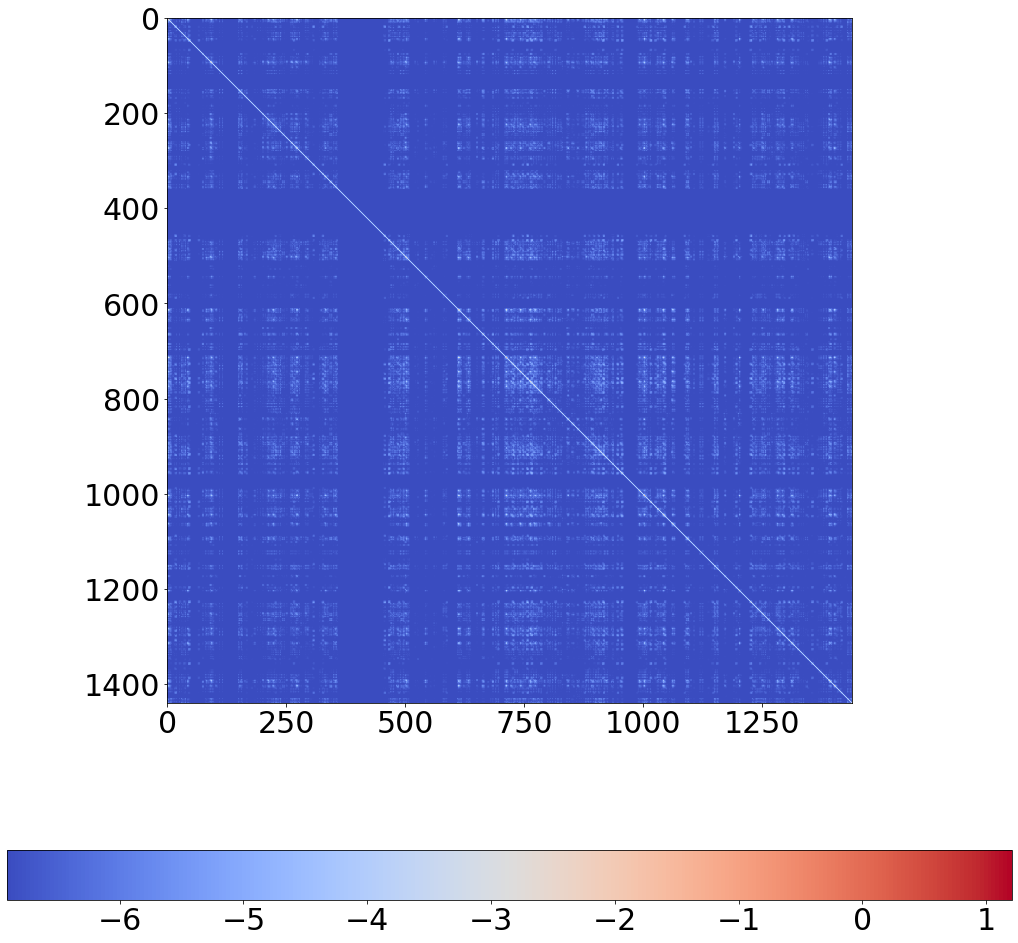}
         \caption{TENGraD}
        \label{fig:TENGraD_fisher_inv_l}
    \end{subfigure}
    \begin{subfigure}[b]{0.25\textwidth}
        \includegraphics[width=\textwidth]{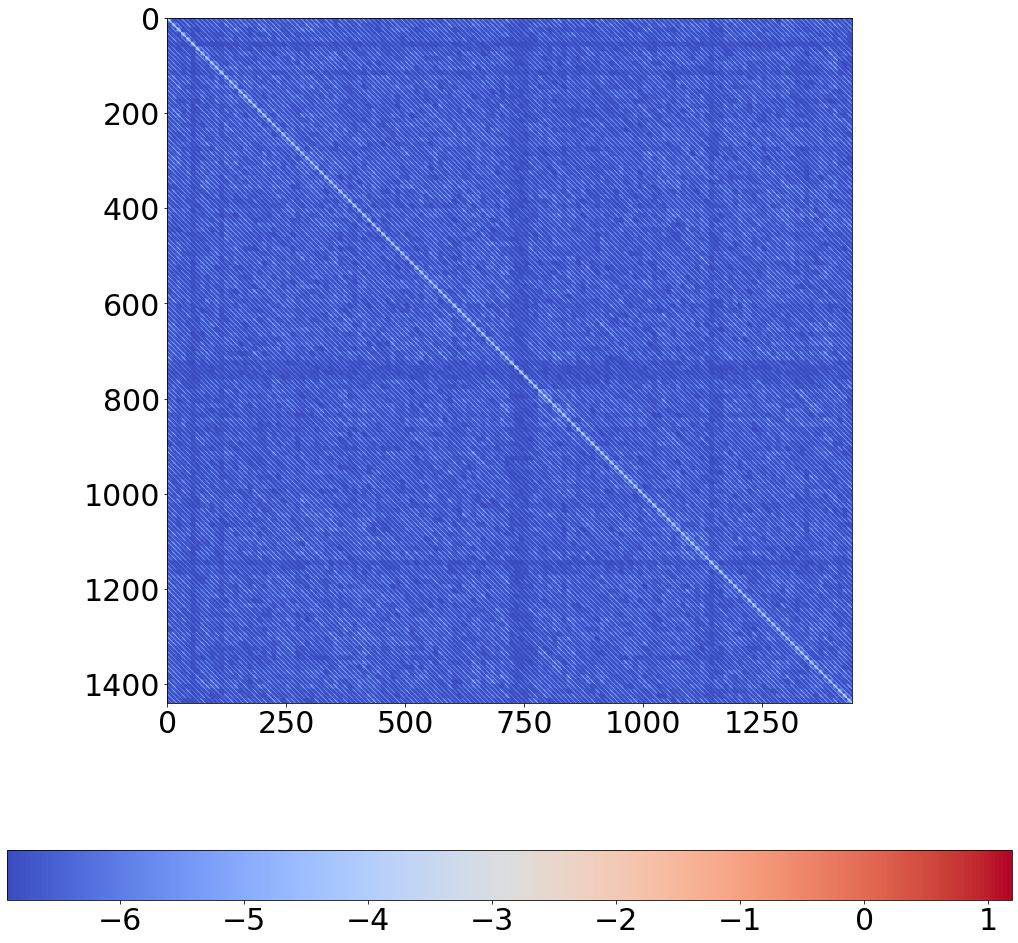}
         \caption{KFAC}
        \label{fig:kfac_fisher_inv_l}
    \end{subfigure}
    \caption{Structure of block inverses for the fully connected layer in 3C1F with Fashion-MNIST.  }\label{fig:approximation_quality_l}
\end{figure*}

In order to compare the quality of FIM approximation, we compare the diagonal block approximations in TENGraD and KFAC with the exact NGD method. The structure for the third convolution layer and the fully connected layer in the 3C1F model are shown in \autoref{fig:approximation_quality_3c} and \autoref{fig:approximation_quality_l}. TENGraD keeps the block structure of the Fisher inverse similar to that of exact NGD while KFAC has a different pattern due to its specific Kronecker structure. Moreover, KFAC still approximates the block inverse with a diagonally dominant matrix in the third convolution layer.

\subsubsection{Sensitivity to Hyperparameters}
\label{sec:supp_sensitivity_hp}
\begin{figure*}[h!]
    \centering
    \begin{subfigure}[b]{0.32\textwidth}
        \includegraphics[width=\textwidth]{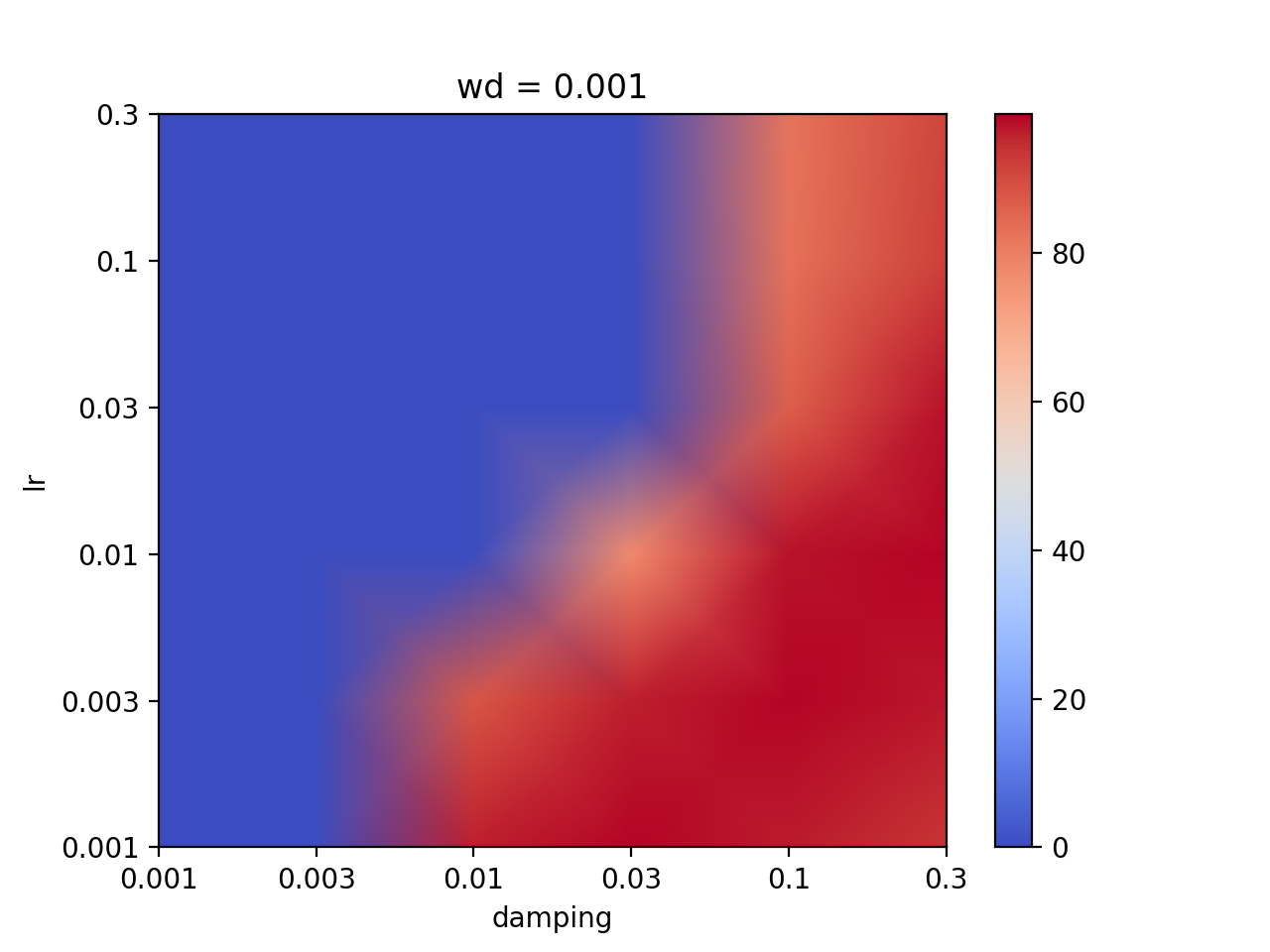}
    \end{subfigure}
    \begin{subfigure}[b]{0.32\textwidth}
        \includegraphics[width=\textwidth]{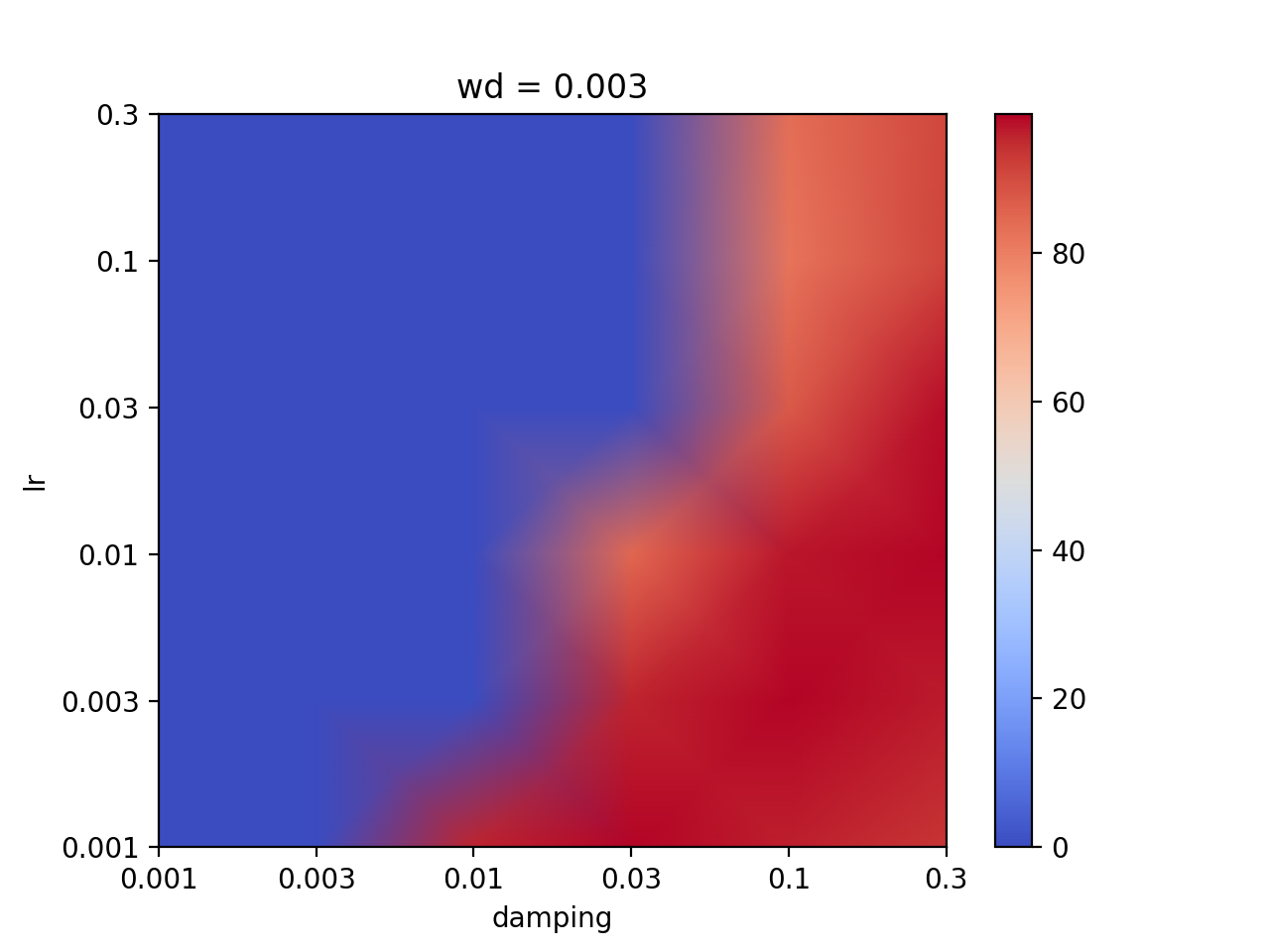}
    \end{subfigure}
    \begin{subfigure}[b]{0.32\textwidth}
        \includegraphics[width=\textwidth]{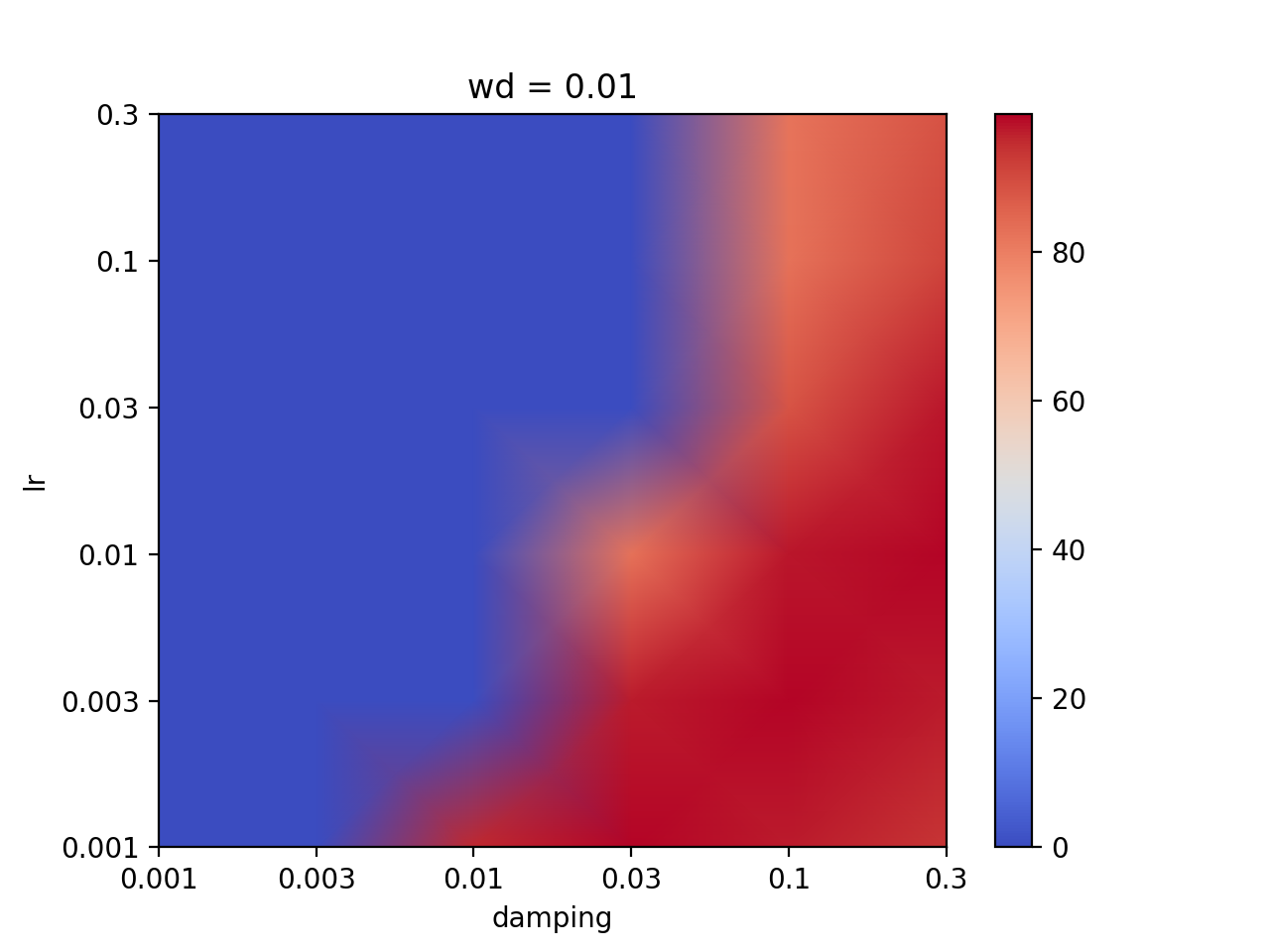}
        %  \caption{Test}
        % \label{fig:hp_test}
    \end{subfigure}
    \caption{The train accuracy for the Fashion-MNIST benchmark under various hyperparameter settings for the test samples. }\label{fig:hyperparameters_train}
\end{figure*}

\begin{figure*}[h!]
    \centering
    \begin{subfigure}[b]{0.32\textwidth}
        \includegraphics[width=\textwidth]{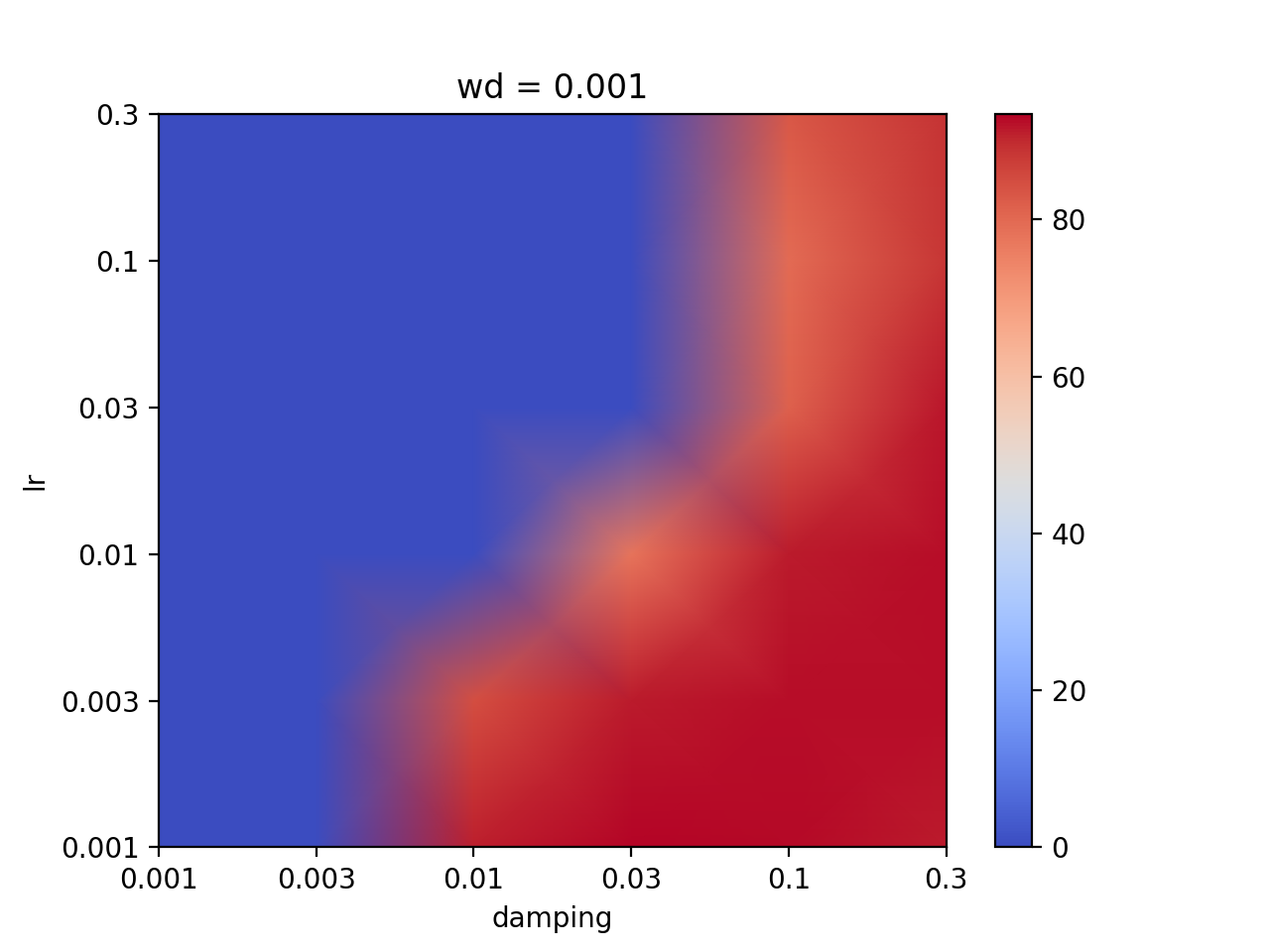}
        %  \caption{a}
        % \label{fig:hp_train_001}
    \end{subfigure}
    \begin{subfigure}[b]{0.32\textwidth}
        \includegraphics[width=\textwidth]{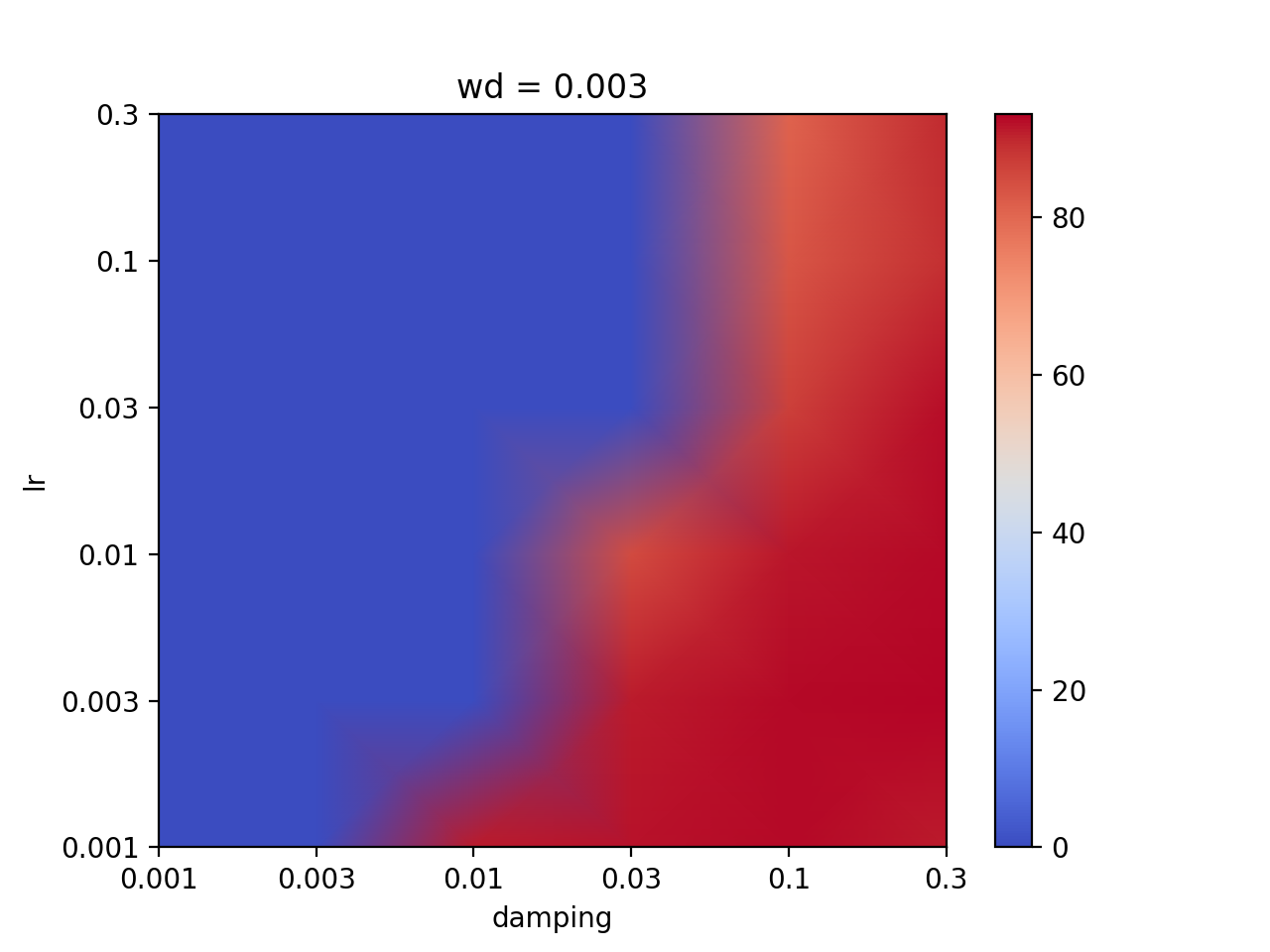}
        %  \caption{b}
        % \label{fig:hp_test}
    \end{subfigure}
    \begin{subfigure}[b]{0.32\textwidth}
        \includegraphics[width=\textwidth]{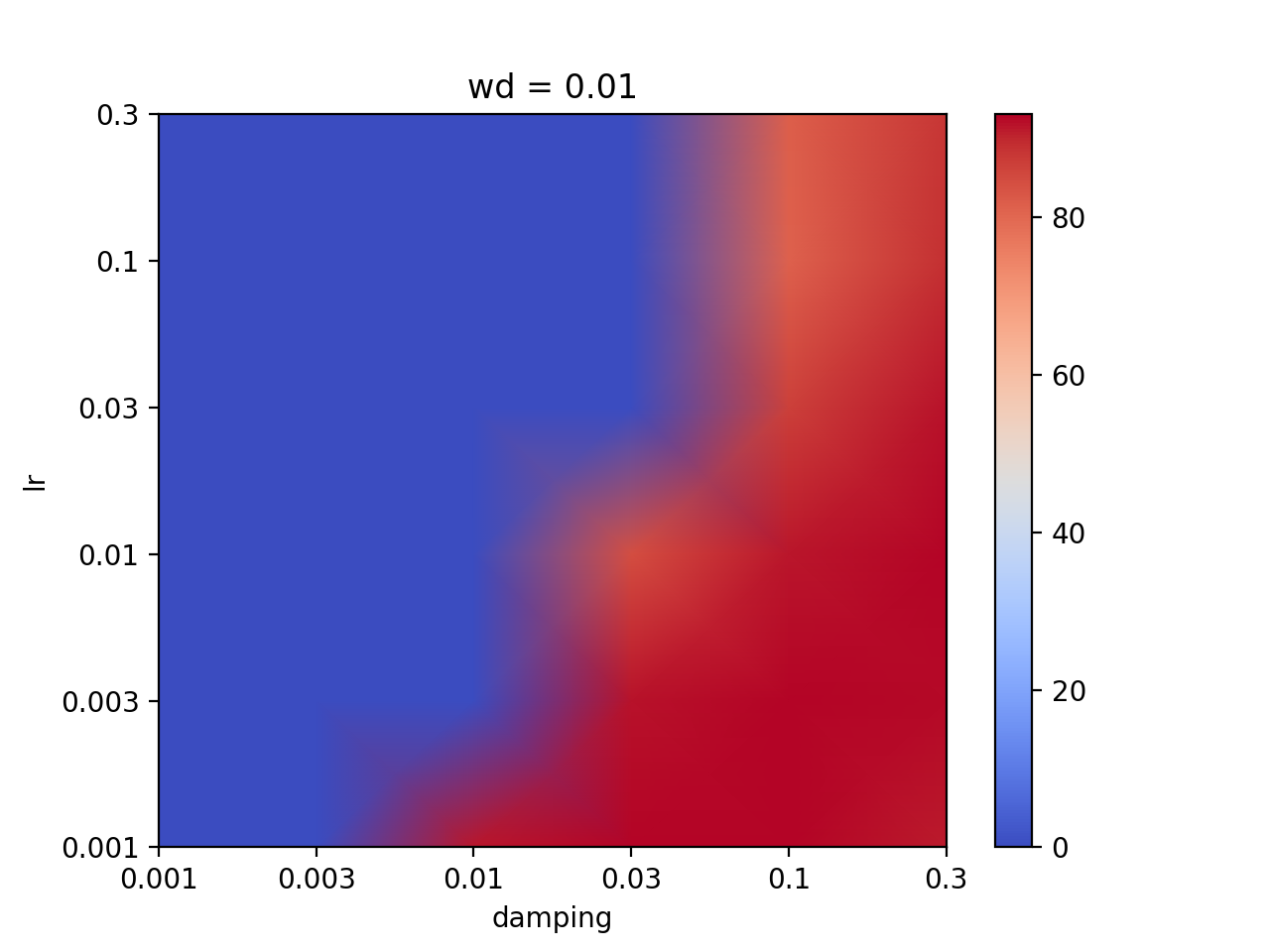}
        %  \caption{Test}
        % \label{fig:hp_test}
    \end{subfigure}
    \caption{The test accuracy for Fashion-MNIST benchmark under various hyperparameter settings for the test samples. }\label{fig:hyperparameters_test}
\end{figure*}

To demonstrate the robustness, we examine the train and test accuracy under various hyperparameter (HP)
settings and show TENGraD is stable under a wide range for the HPs. The train and test accuracy for the Fashion-MNIST benchmark under different HP settings is shown in \autoref{fig:hyperparameters_train} and \autoref{fig:hyperparameters_test}. Also, we observe that a larger damping parameter allows for larger step sizes. TENGraD allows for a wide range of HP to be used with almost no loss of accuracy on both train and test data.

\subsection{Error Bar Plots}
\label{sec:err_bars}
\begin{figure*}[h!]
    \centering
    \begin{subfigure}[b]{0.47\textwidth}
        \includegraphics[width=\textwidth]{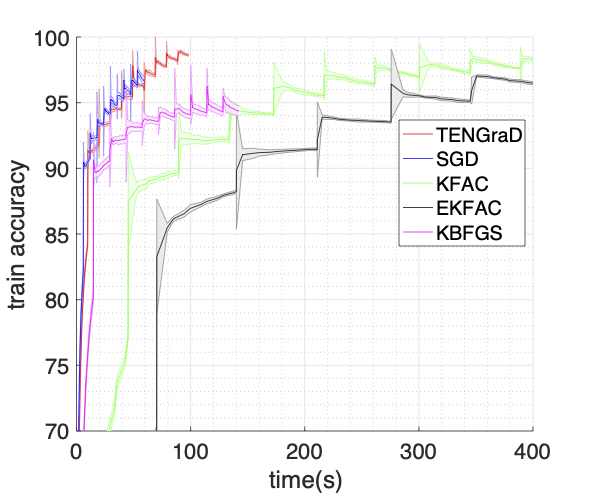}
         \caption{Train}
        % \label{fig:training_fashion}
    \end{subfigure}
    \begin{subfigure}[b]{0.47\textwidth}
        \includegraphics[width=\textwidth]{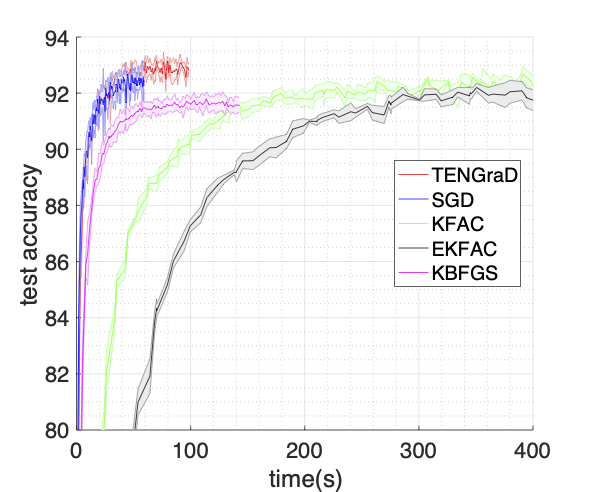}
         \caption{Test}
    \end{subfigure}
    \caption{ The error bars of train and test accuracy for Fashion-MNIST benchmark.}\label{fig:err_fashion}
\end{figure*}

\begin{figure*}[h!]
    \centering
    \begin{subfigure}[b]{0.32\textwidth}
        \includegraphics[width=\textwidth]{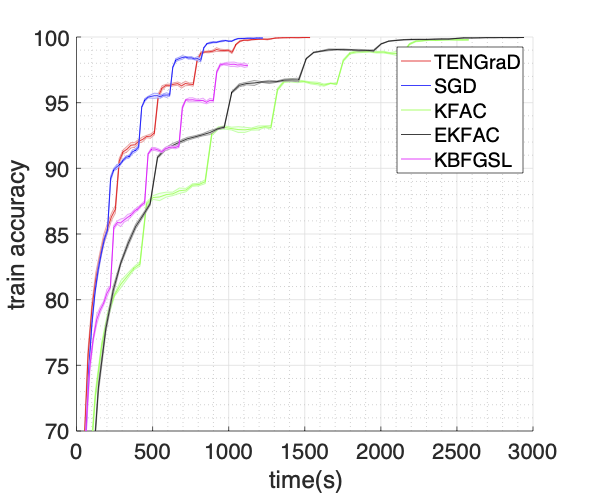}
         \caption{WideResNet}
        % \label{fig:training_fashion}
    \end{subfigure}
    \begin{subfigure}[b]{0.32\textwidth}
        \includegraphics[width=\textwidth]{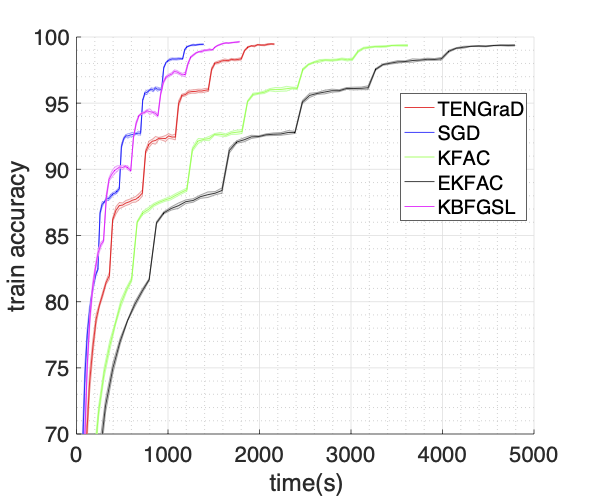}
         \caption{MobileNetV2}
        \label{fig:err_c10_train_mn}
    \end{subfigure}
    \begin{subfigure}[b]{0.32\textwidth}
        \includegraphics[width=\textwidth]{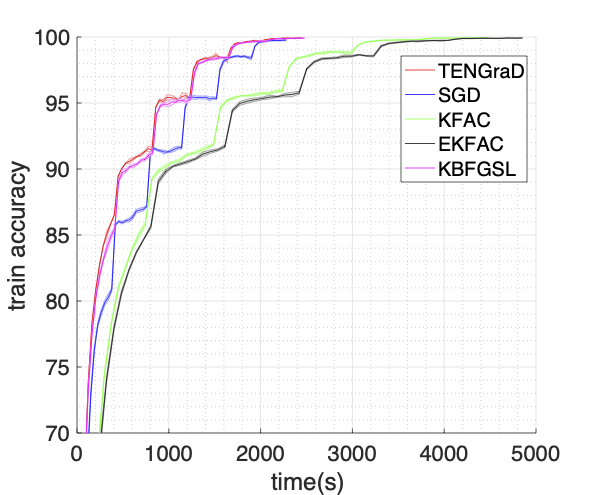}
         \caption{DenseNet}
        % \label{fig:training_cifar100}
    \end{subfigure}
    \caption{ The error bars of train accuracy for CIFAR-10 benchmark.}\label{fig:err_c10_train}
\end{figure*}

\begin{figure*}[h!]
    \centering
    \begin{subfigure}[b]{0.32\textwidth}
        \includegraphics[width=\textwidth]{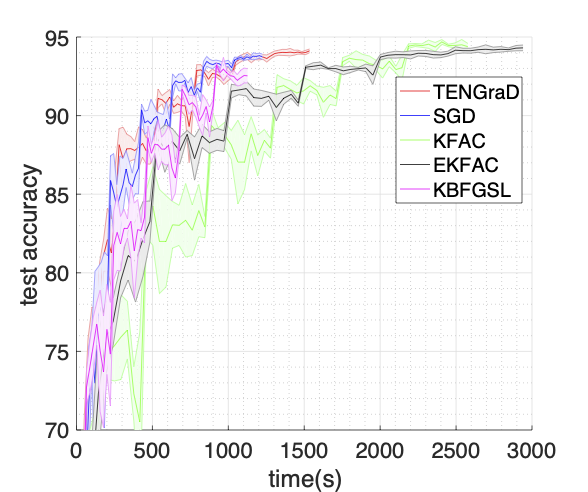}
         \caption{WideResNet}
        % \label{fig:training_fashion}
    \end{subfigure}
    \begin{subfigure}[b]{0.32\textwidth}
        \includegraphics[width=\textwidth]{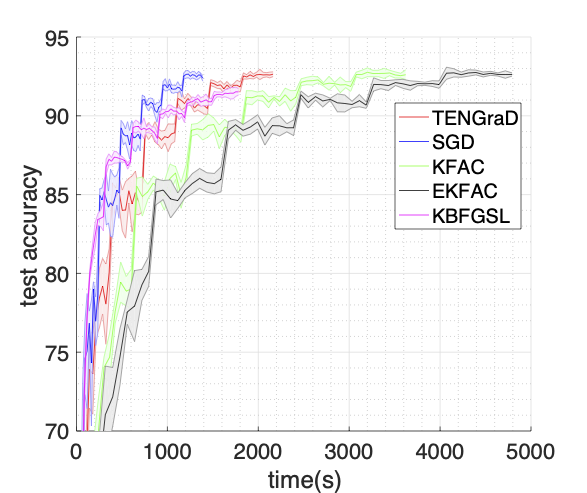}
         \caption{MobileNetV2}
        \label{fig:err_c10_test_mn}
    \end{subfigure}
    \begin{subfigure}[b]{0.32\textwidth}
        \includegraphics[width=\textwidth]{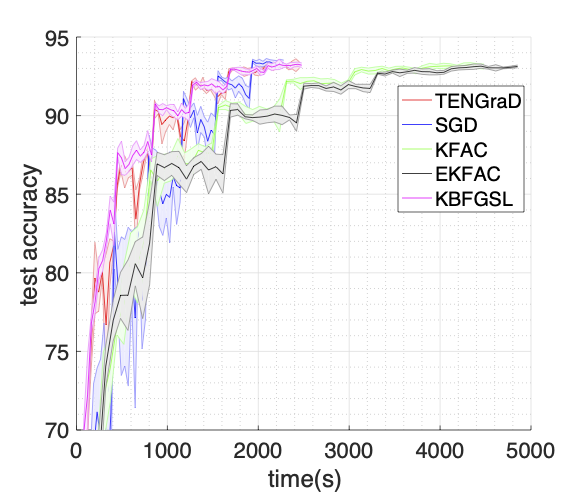}
         \caption{DenseNet}
        % \label{fig:training_cifar100}
    \end{subfigure}
    \caption{ The error bars of test accuracy for CIFAR-10 benchmark.}\label{fig:err_c10_test}
\end{figure*}

\begin{figure*}[h!]
    \centering
    \begin{subfigure}[b]{0.32\textwidth}
        \includegraphics[width=\textwidth]{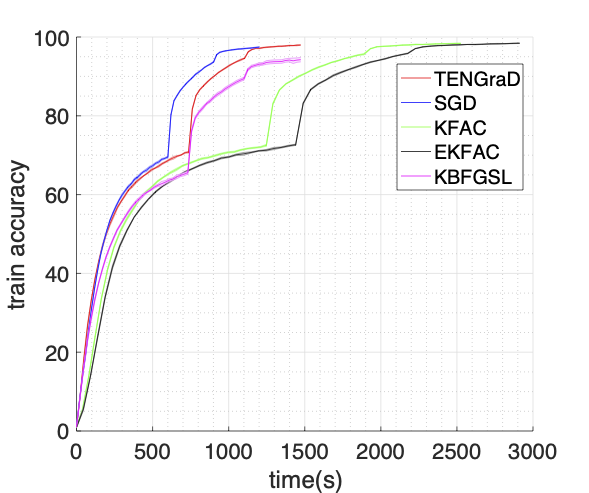}
         \caption{WideResNet}
        \label{fig:err_c100_train_dn}
    \end{subfigure}
    \begin{subfigure}[b]{0.32\textwidth}
        \includegraphics[width=\textwidth]{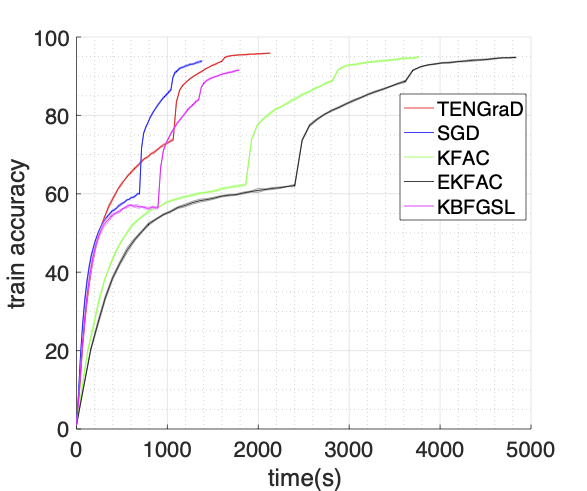}
         \caption{MobileNetV2}
        % \label{fig:err_c10_train_mn}
    \end{subfigure}
    \begin{subfigure}[b]{0.32\textwidth}
        \includegraphics[width=\textwidth]{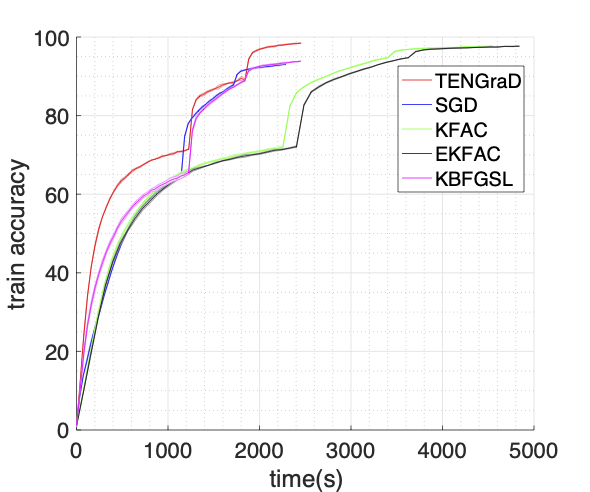}
         \caption{DenseNet}
        % \label{fig:training_cifar100}
    \end{subfigure}
    \caption{ The error bars of train accuracy for CIFAR-100 benchmark.}\label{fig:err_c100_train}
\end{figure*}

\begin{figure*}[h!]
    \centering
    \begin{subfigure}[b]{0.32\textwidth}
        \includegraphics[width=\textwidth]{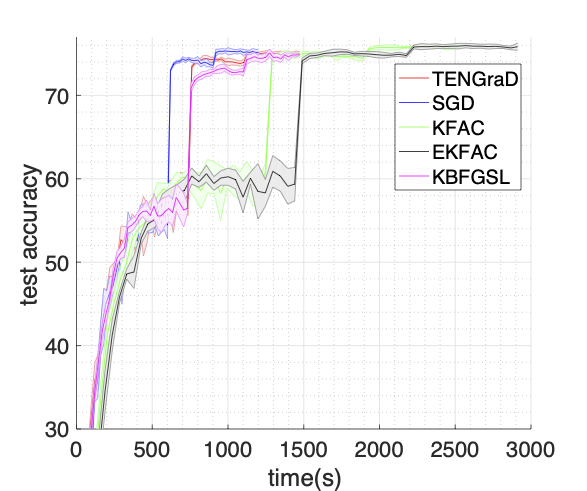}
         \caption{WideResNet}
        \label{fig:err_c100_test_dn}
    \end{subfigure}
    \begin{subfigure}[b]{0.32\textwidth}
        \includegraphics[width=\textwidth]{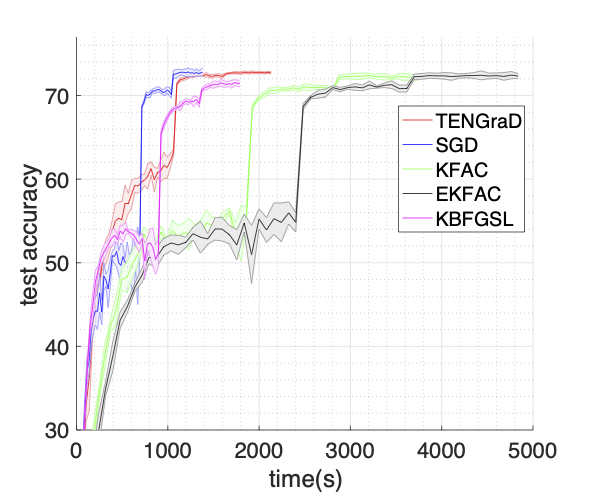}
         \caption{MobileNetV2}
        % \label{fig:err_c10_test_mn}
    \end{subfigure}
    \begin{subfigure}[b]{0.32\textwidth}
        \includegraphics[width=\textwidth]{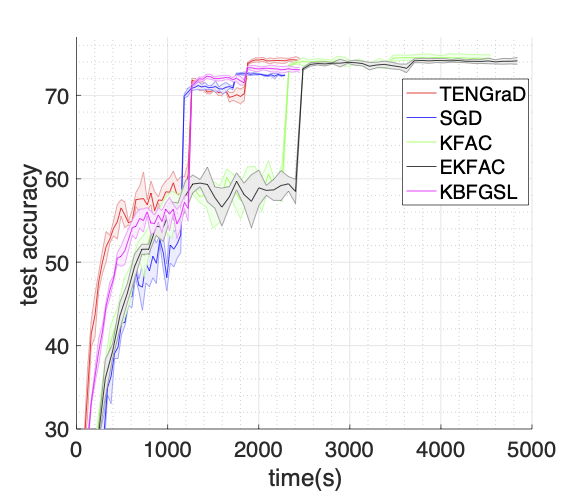}
         \caption{DenseNet}
        % \label{fig:training_cifar100}
    \end{subfigure}
    \caption{ The error bars of test accuracy for CIFAR-100 benchmark.}\label{fig:err_c100_test}
\end{figure*}

The error bars for the Fashion-MNIST benchmark are shown in \autoref{fig:err_fashion}. TENGraD outperforms both SGD and other NGD methods in time and accuracy.
The error bar plots for five different runs of the CIFAR-10 benchmark are shown in \autoref{fig:err_c10_train} and \autoref{fig:err_c10_test}. TENGraD outperforms other NGD methods in time and accuracy for both train and test. In \autoref{fig:err_c10_train_mn} the train accuracy of KBFGS appears to be better than that of TENGraD but the corresponding test accuracy in \autoref{fig:err_c10_test_mn} shows that TENGraD has a higher test accuracy and does not over-fit to data. The error bars of train and test for the CIFAR-100 benchmark are shown in \autoref{fig:err_c100_train} and \autoref{fig:err_c100_test}. TENGraD outperforms SGD in time and accuracy in training the DenseNet model as shown in \autoref{fig:err_c100_train_dn} and achieves a better accuracy compared to all other methods as shown in \autoref{fig:err_c100_test_dn}. TENGraD is also competitive with SGD on other benchmarks and outperforms other NGD methods.

%%%%%%%%%%% CIFAR-100 ERROR BARS

\end{document}